\documentclass[10pt,final,twocolumn]{IEEEtran}
\usepackage[english]{babel}
\usepackage[margin=0.95in]{geometry}
\usepackage{cite}
\usepackage{algpseudocode}
\usepackage{graphicx}
\usepackage{amsmath}
\usepackage{algpseudocode}
\usepackage{algorithm}
\usepackage{caption}
\usepackage{subcaption}
\usepackage{float}
\usepackage{ifthen}
\usepackage{multicol}
\usepackage{optidef}
\usepackage{amsmath}
\usepackage{comment}
\usepackage{xcolor}
\usepackage{amssymb}
\usepackage{multirow}
\usepackage{nicefrac}
\usepackage{hyperref}
\usepackage{mathtools}
\usepackage[english]{babel}
\usepackage{amsthm}

\newtheorem{assumption}{Assumption}

\DeclareRobustCommand{\rchi}{{\mathpalette\irchi\relax}}
\newcommand{\irchi}[2]{\raisebox{\depth}{$#1\chi$}}
\newtheorem{theorem}{Theorem}

\begin{document}
\title{Sparse Training for Federated Learning with Regularized Error Correction}

\author{Ran Greidi, Kobi Cohen (\emph{Senior Member, IEEE})
\thanks{
		Ran Greidi and K. Cohen are with the School of Electrical and Computer Engineering, Ben-Gurion University of the Negev, Beer-Sheva, Israel (e-mail:rangrei@post.bgu.ac.il; yakovsec@bgu.ac.il).
	}
	\thanks{This work was supported by the Israel Science Foundation under Grant 2640/20. 
}
\thanks{Open-source code for the algorithm and simulations developed in this paper can be found on GitHub at \cite{Greidi_FLARE_Federated_Learning_2023}.}
\thanks{This work has been submitted to the IEEE for possible publication.
Copyright may be transferred without notice, after which this version
may no longer be accessible.
}}

\maketitle

\begin{abstract}
Federated Learning (FL) is an emerging paradigm that allows for decentralized machine learning (ML), where multiple models are collaboratively trained in a privacy-preserving manner. It has attracted much interest due to the significant advantages it brings to training deep neural network (DNN) models, particularly in terms of prioritizing privacy and enhancing the efficiency of communication resources when local data is stored at the edge devices. However, since communications and computation resources are limited, training DNN models in FL systems face challenges such as elevated computational and communication costs in complex tasks.

Sparse training schemes gain increasing attention in order to scale down the dimensionality of each client (i.e., node) transmission. Specifically, sparsification with error correction methods is a promising technique, where only important updates are sent to the parameter server (PS) and the rest are accumulated locally. While error correction methods have shown to achieve a significant sparsification level of the client-to-PS message without harming convergence, pushing sparsity further remains unresolved due to the staleness effect. In this paper, we propose a novel algorithm, dubbed Federated Learning with Accumulated Regularized Embeddings (FLARE), to overcome this challenge. FLARE presents a novel sparse training approach via accumulated pulling of the updated models with regularization on the embeddings in the FL process, providing a powerful solution to the staleness effect, and pushing sparsity to an exceptional level. Our theoretical analysis demonstrates that FLARE's regularized error feedback achieves significant improvements in scalability with sparsity parameter. The empirical performance of FLARE is validated through extensive experiments on diverse and complex models, achieving a remarkable sparsity level (10 times and more beyond the current state-of-the-art) along with significantly improved accuracy. Additionally, an open-source software package has been developed for the benefit of researchers and developers in related fields.
\end{abstract}


\section{Introduction}

The rapid expansion of 5G and IoT applications and the swift evolution of edge devices capabilities in the recent years enabled ML based on a centralized data centers to become distributed. Recent growth of technologies empowered edge devices with computational capabilities, made it possible to perform ML tasks locally. The centralized to distributed transformation has eliminated the need of transferring data to a centralized data centers which imposes significant strain on network resources and exposes private and sensitive data. FL is an emerging ML crafted for training models across numerous clients, each containing local datasets, all achieved without the necessity for a direct exchange of communication-costly sensitive data with a central PS\cite{mcmahan2017communication, aledhari2020federated, Gafni_2022}.

FL is particularly well-suited for mobile applications, such as those in 5G, IoT, and cognitive radio systems. This suitability stems from privacy concerns related to local data stored at edge devices \cite{mcmahan2017communication, aledhari2020federated, amiri2020machine}. Additionally, the communication aspects of FL have been a subject of exploration in recent years, spanning both digital \cite{chen2020joint, abad2019hierarchical, naparstek2018deep, gafni2022distributed, gafni2022learning, ami2023client, salgia2023communicationefficient} and analog \cite{Sery_2020, Sery_2021, 9562537, 10168898} communications. In practical scenarios, the effective implementation of FL encounters two challenges: The communication bottleneck, denoting the burden on the communication channel due to uplink transmissions by all devices (clients) to the PS, and the computational constraints of resource-constrained devices. Addressing these issues is a dynamic and evolving field with numerous prior works spanning various areas, including weight pruning\cite{han2016deep, li2017pruning, livne2020pops}, over-the-air methods \cite{10168898, zhu2020one, Sery_2021, Sery_2020, 9562537}, and model or gradient compression\cite{ZHAO20238669,doi:10.1137/21M1450677, chen2021joint, li2022soteriafl, NEURIPS2022_cd86c6a8, xue2023riemannian, li2023convergence, Rothchild20208253,8889996, wang2023communication}. Popular approaches and techniques that have been explored to reduce the communication bottleneck issue can be mainly grouped into quantization and sparsification. While quantization techniques reduce the size of transmitted model updates through lower precision representations, thereby reducing communication overhead\cite{seide2014-bit}, sparsification methods aim to reduce communication overhead efficiently by restricting the updates transmitted by each client and aggregated by the server into a smaller subset.
    
Among sparsification methods, Top-$K$ is a commonly adopted sparse training scheme. In the Top-$K$ method, each client receives a global model, performs local optimization, and then transmits only the gradients or the model deltas corresponding to the Top-$K$ absolute values. Performing the Top-$K$ method on gradients is suitable for distributed gradient descent algorithms such as FedSGD\cite{mcmahan2017communication}, but restricts each client for only one optimization step. Thus, it is communication-inefficient due to the increased communication overhead caused by more frequent transmissions. To allow for communication-efficient FL with more local steps (i.e. FedAvg\cite{mcmahan2017communication}), Top-$K$ can be executed via model deltas. Model deltas signify the individual parameter changes of each client with respect to the global model (i.e. subtracting the global model from the new local model), subsequent to local optimization at each round. With the server retaining the global model for the ongoing round, it can reconstruct the new global model for the subsequent round by aggregating sparse updates from all clients. This occurs after each client has conducted Top-$K$ on its model delta\cite{Gafni_2022}. 
    
To enhance the performance of Top-$K$ sparsification, Gradients Correction methods have been proposed in \cite{Strom2015ScalableDD}, \cite{aji-heafield-2017-sparse}. These approaches selectively transmit only the top gradients, while locally accumulating the remaining ones, collecting all unsent gradients into residuals. As these residuals accumulate to a sufficient magnitude, all gradients are effectively transmitted. Remarkably, these methods have achieved a sparsification level of $99.9\%$ without causing significant convergence damage. However, subsequent observations in \cite{DBLP:journals/corr/abs-1712-01887} revealed residual damage due to the staleness effect, a consequence of delayed and outdated updates caused by accumulation. The authors addressed the staleness effect by employing momentum factor masking and warm-up training. In another effort to mitigate the impact of stale updates, a new framework was proposed in \cite{9148987}, leveraging batch normalization and optimizer adjustment, demonstrating effectiveness with $99.9\%$ sparsity and improved convergence. Nevertheless, these techniques rely on gradients manipulations, rendering them impractical for FL with more than a single optimization step. Incorporating the concept of Gradients Correction, which involves accumulating residuals, into FL with more than one optimization step, \cite{DBLP:journals/corr/abs-1805-08768} introduced an Error Correction method. The same principle can be applied to model deltas, thereby reducing communication overhead through less frequent parameter exchanges with the PS. This method successfully achieved a sparsity level of $99.9\%$ with less frequent communication. 

Although recent error correction methods have demonstrated notable success in achieving a substantial sparsification level of the client-to-PS message without causing significant harm to convergence, as discussed above, further advancing sparsity remains an unresolved challenge due to the staleness effect. In this paper, our aim is to tackle this issue. We will demonstrate that our proposed novel method achieves an outstanding level of sparsity (a magnitude 100 times beyond existing sparsification methods), while simultaneously preserving accuracy performance.

\subsection{Main Results}

We develop a novel FL algorithm with low-dimensional embeddings through model transmission sparsification for communication-efficient learning, dubbed Federated Learning with Accumulated Regularized Embeddings (FLARE). FLARE employs a sophisticated Top-$K$ sparsification and error accumulation FL method, significantly reducing communication costs in FL systems compared to existing methods. Our motivation stems from addressing the staleness effect, identified as a fundamental reason for the failure of Error Correction techniques when sparsity is pushed to extreme\cite{9148987,DBLP:journals/corr/abs-1712-01887}. This root cause is a critical issue hindering the convergence of the FL process when using error accumulation techniques. To overcome this limtation, FLARE is designed based on a novel Error Correctoins approach with regularized embeddings. 

The Error Correction for model deltas is implemented as follows: During each iterative round of the FL process, after the client completes the local model update procedure, the client selectively identifies and transmits only those model parameter deltas deemed as the Top-$K$ in magnitude, determined by their absolute values. These selective updates are then transmitted to the PS for aggregation. Additionally, clients accumulate residuals locally, acting as corrections to counteract the Top-$K$ sparsification. That is, residuals initially omitted from the Top-$K$ selection are accumulated locally. Over time, these residuals become large enough to be transmitted, ensuring that all changes are eventually sent.

The FLARE algorithm, proposed in this study, introduces an innovative sparsification approach with error accumulation for transmitting a low-dimensional representation of the model at each iteration. It employs a unique technique that modifies the objective loss without necessitating intensive communication or computational resources for updating the entire high-dimensional model. The accumulator at each client is dedicated solely to storing weight updates that were not transmitted during training, compensating for sparse communication by transmitting delayed updates once they accumulate sufficiently. As a result, each client retains locally stored accumulator data containing valuable yet unused optimization information. This information is leveraged during the in-round optimization steps. Additionally, we capitalize on these values for each client by introducing a new and client-specific loss term during each communication round. This term is crafted to both minimize the client objective loss and address the staleness effect. It is utilized to regularize the weight updates concerning weights that were not transmitted, thereby adjusting the optimization trajectory of each client closer to its original uncompressed track. For a detailed description of the FLARE algorithm, please refer to Section \ref{sec:FLARE}.

Second, for performance analysis, we conducted both theoretical and empirical evaluations to assess the efficiency of FLARE. The theoretical convergence analysis demonstrates that FLARE's regularized error feedback not only matches state-of-the-art techniques in terms of convergence rate with time, but also achieves significant improvements in scalability with respect to the sparsity parameter.
In terms of empirical evaluations, we conducted extensive simulation experiments involving various ML tasks of different sizes. The experiments utilized six distinct models for different FL settings, namely Fully Connected (FC), Convolutional Neural Network (CNN), VGG 11, VGG 16, VGG 19 and GRU models, applied to MNIST, CIFAR10, and "The Complete Works of William Shakespeare" datasets. The results affirm FLARE's superior performance compared to existing methods, demonstrating higher accuracy and sparsity levels. Specifically, FLARE achieves a sparsity level of $99.999\%$, surpassing the $99.9\%$ sparsity level of existing methods and the $99.99\%$ achieved by the current state-of-the-art. This represents a magnitude 10 times and more beyond the current state-of-the-art, accompanied by significantly improved accuracy. This remarkable advancement enables FL in bandwidth-limited networks. The robustness and efficacy of FLARE are underscored by these results, marking a significant progress in the field. The simulation results are detailed in Section \ref{sec:experiments}.

Third, we have developed an open-source implementation of the FLARE algorithm using the TensorFlowFederated API. The implementation is available on GitHub at \cite{Greidi_FLARE_Federated_Learning_2023}. Our experimental study highlights the software's versatility across various challenging environments. We actively encourage researchers and developers in related fields to utilize this open-source software.

\section{System Model and Problem Statement} 

We consider an FL system comprised of a parameter server (PS) and $N$ clients. The dataset in the FL system $\chi$ is distributed among the $N$ clients, each of which holds its local dataset $\chi^i \subset \chi$, $i=1,...,N$. Ideally, the objective is to minimize the global loss, which is defined, for given weights $\bar{w} \in R^d $ by: 
\begin{equation}
\label{eq:erm}
F(\bar{w};\chi) = \frac{1}{|\chi|} \sum_{x \in \chi} f(x,\bar{w}).
\end{equation}
The minimizer of \eqref{eq:erm} is known as the empirical risk minimizer.

In FL systems, each client holds its distinct and private dataset, denoted as $\chi^i$, which the PS does not directly access. This setup prevents us from directly solving \eqref{eq:erm} in a centralized manner. Therefore, the optimization process in FL systems involves utilizing the computational resources of the distributed clients to minimize iteratively the local loss function defined as follows:
\begin{equation}
\label{eq:local_opt}
F^i(\bar{w};\chi^i) = \frac{1}{|\chi^i|} \sum_{x \in \chi^i} f(x,\bar{w}).
\end{equation}
At each iteration round $k$, each client receives from the PS a global model denoted by $\bar{w}_k \in R^d$. Then, it minimizes its local loss $F^i(\bar{w}_k;\chi^i)$ using its own dataset $\chi^i$ to obtain $\bar{w}_{k}^i$, and transmits its updated local weights $\bar{w}^{i}_{k}$ to the PS. The PS aggregates the updated model weights received from all clients, and broadcasts an updated global model weights $\bar{w}_{k+1}$ back to the clients for the next iteration. 

In this paper, we aim to develop an FL algorithm that reduces the dimensionality of the transmitted model at each iteration, such that the procedure of the FL process approaches the performance of the global minimizer \ref{eq:erm}, all while incurring minimal communication costs in terms of total transmission data.\vspace{0.2cm}

\noindent
\textbf{Notations:} Throughout this paper, superscripts refer to the client index, while subscripts denote the iteration round index. Please refer to Table~\ref{table:notation} for a comprehensive list of notations used in this paper.
    
    \begin{table} [htbp!]
    \centering
    \caption{Notations}
    
    \label{table:notation}
    \begin{center}
    \begin{tabular}{ |p{0.9cm}||p{4.5cm}|}
    \hline
     Notation& Description \\
    \hline \hline
    $N$ & Number  of clients \\ 
    \hline
    $d$ & Weight dimension  \\
    \hline
    $T$ & Number of rounds \\ 
    \hline
    $k$ & Round index  \\
    \hline      
    $\chi^i$ & Local dataset of client $i$\\
    \hline    
    $\chi$ & Global dataset \\
    \hline
    $F_i$ &  Local empirical loss for client $i$\\
    \hline
    $F$ &   Global empirical loss\\
    \hline    
    $f$ &  Objective loss function\\
    \hline
    $\bar{w}^i_k$ &  Client's $i$ model vector at round $k$  \\          \hline 
    $\bar{w}_k$ &  Global model vector at the PS at round $k$ \\
    \hline    
    $\bar{A}_k^i$ &  Accumulator vector for client $i$ at round $k$\\
    \hline
    $E$ &  Number of optimization steps  \\           
    \hline
    $B$ &  Batch size  \\           
    \hline    
    $R,\delta$ &  Sparsity percent, sparsity ratio \\           
    \hline   
    $\gamma$ & Learning rate \\
    \hline
    $\tau,\tau_k$ & FLARE regularization coefficient \\
    \hline
    $a_0$ & FLARE masking coefficient \\
    \hline
    $c$ & decay constant of $\tau$ \\    
    \hline
    $p$ & Number of steps for FLARE objective \\    
    \hline    
    \end{tabular}
    \end{center}  
    \end{table}

\section{The Federated Learning with Accumulated Regularized Embeddings (FLARE) Algorithm}    
\label{sec:FLARE}
    \begin{figure*}
             \centering
             \begin{subfigure}{0.35\textwidth}
                \includegraphics[width=\textwidth]{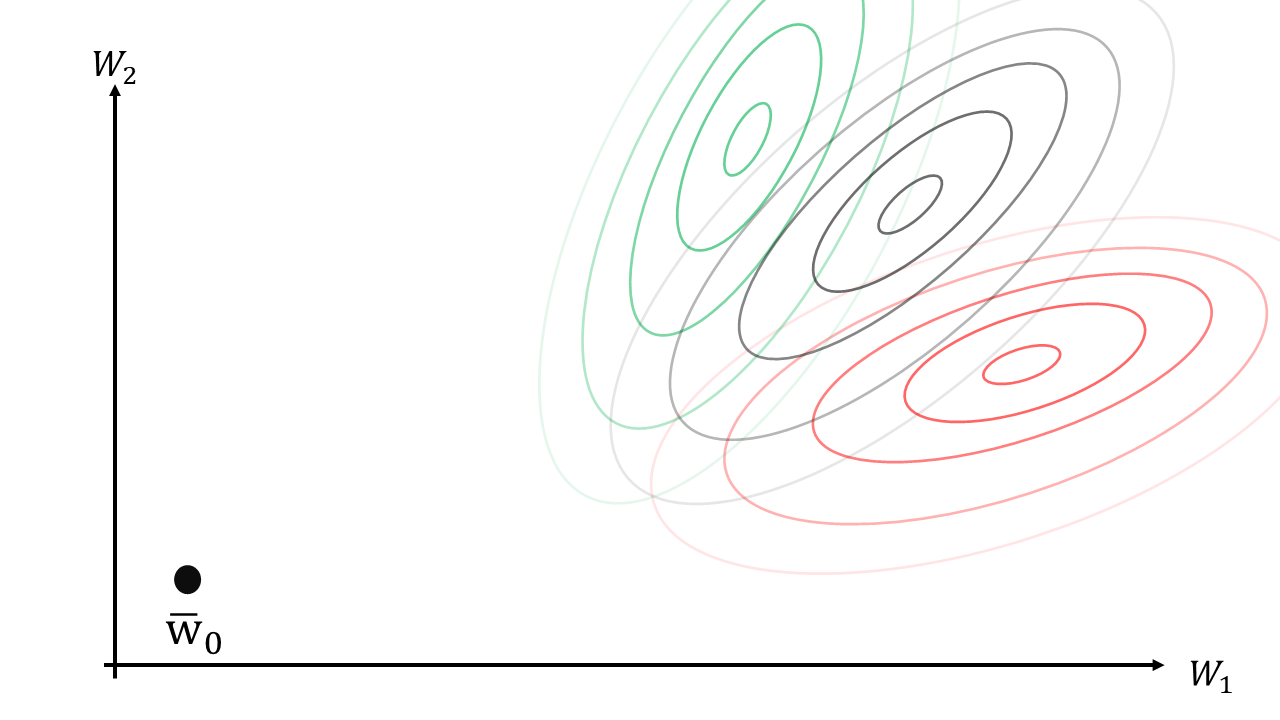}
                \caption{Stage 1}
                \label{fig:myREF_walkthrough/Slide1}
             \end{subfigure}
             \hfill
             \begin{subfigure}{0.35\textwidth}
                \includegraphics[width=\textwidth]{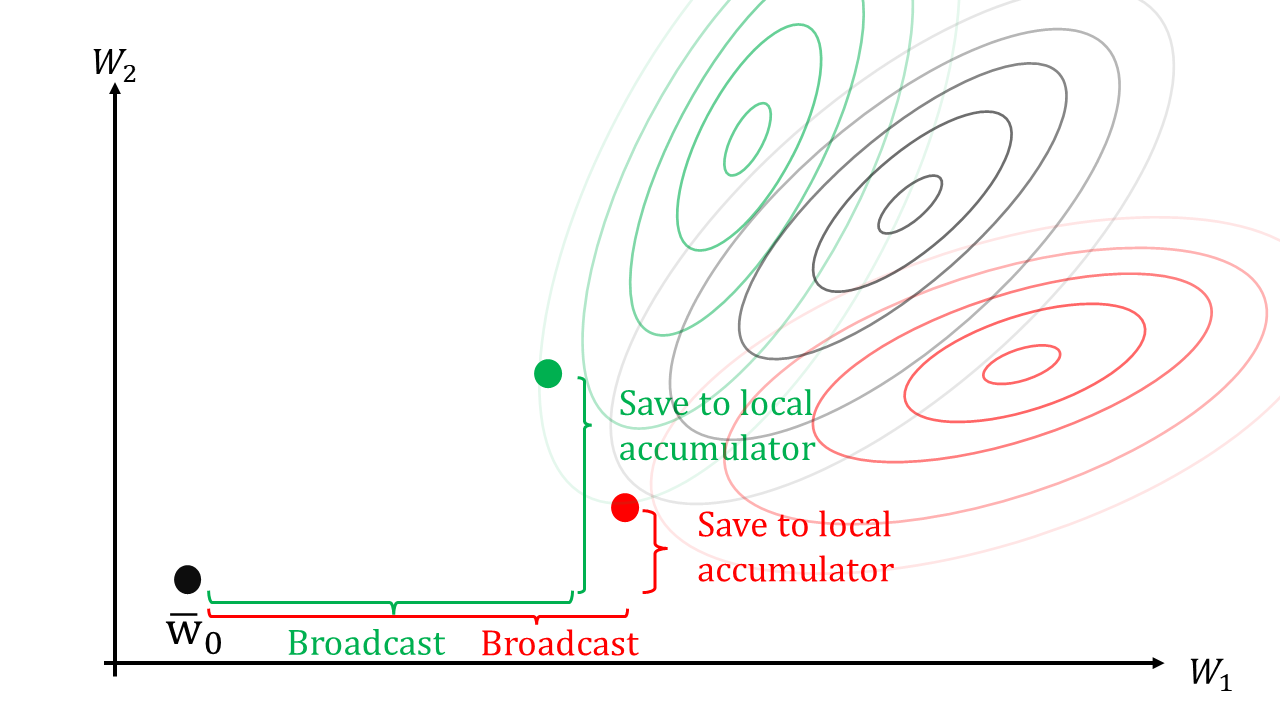}
                \caption{Stage 2}
                \label{fig:myREF_walkthrough/Slide2}
             \end{subfigure}
             \hfill         
             \begin{subfigure}{0.35\textwidth}
                \includegraphics[width=\textwidth]{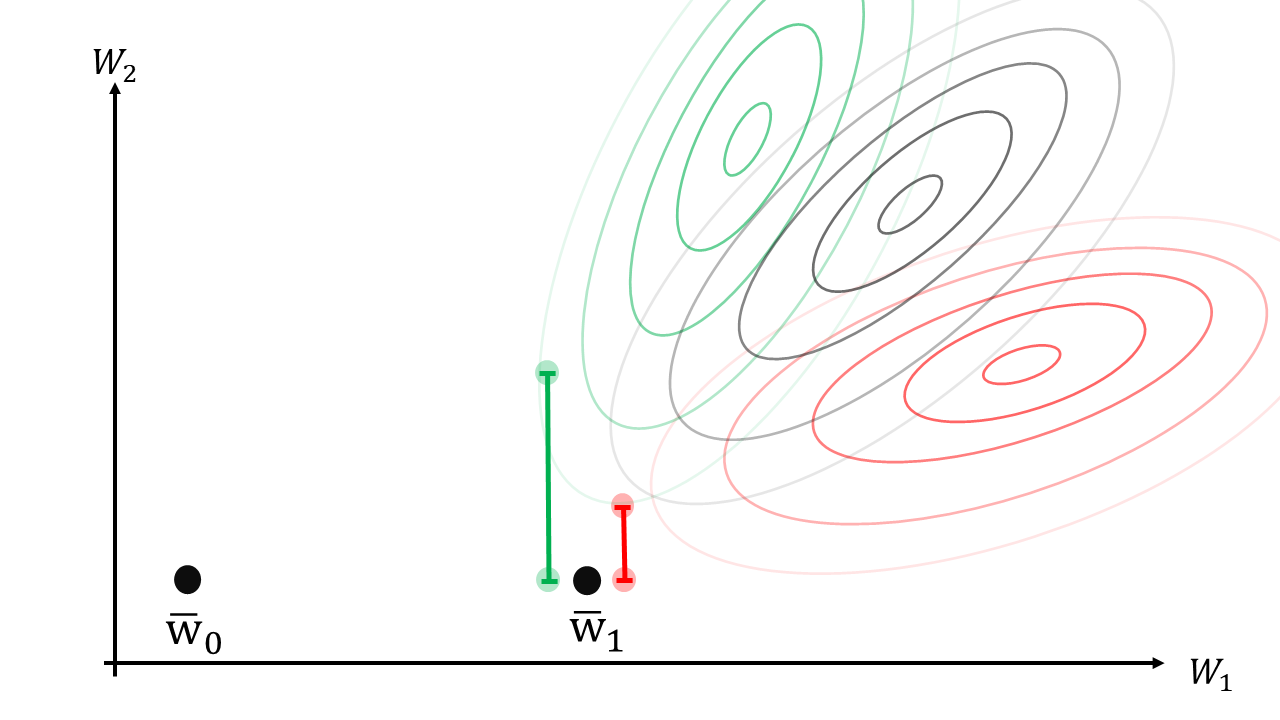}
                \caption{Stage 3}
                \label{fig:myREF_walkthrough/Slide3}
             \end{subfigure}
             \hfill
             \begin{subfigure}{0.35\textwidth}
                \includegraphics[width=\textwidth]{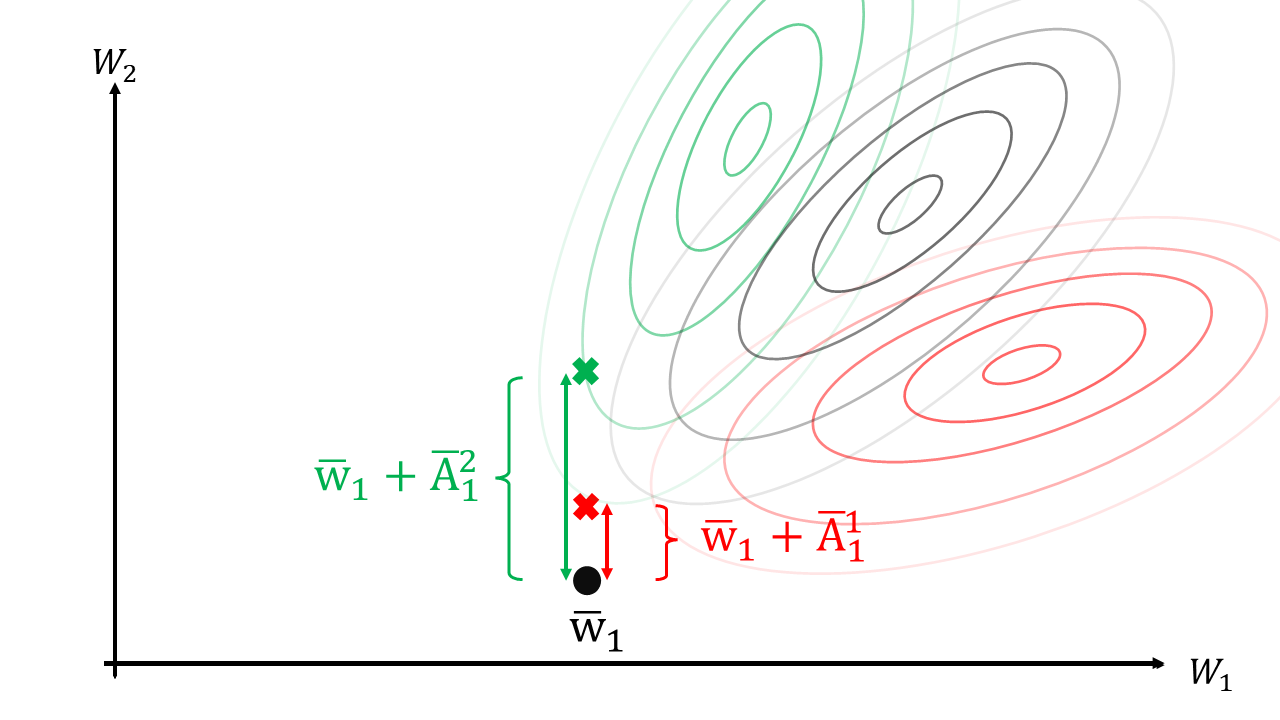}
                \caption{Stage 4}
                \label{fig:myREF_walkthrough/Slide4}
             \end{subfigure}     
        \caption{An illustration of the FLARE algorithm in four stages (refer to a detailed description in Section \ref{ssec:algorithm}): First, the PS broadcasts a global model $\bar{w}_{0}$ to the clients (Stage 1). Subsequently, each client generates a new model, sends its Top-$K$ deltas to the PS, and accumulates the error locally (Stage 2). Next, the PS aggregates all received models and broadcasts a new global model to all clients (Stage 3). FLARE attempts to minimize the staled updates by  minimizing $\tau||\bar{w}-(\bar{w}_{1}+\bar{A}^i_1)||_1$ held by each client (Stage 4). The clients redefine their loss according to \eqref{eqn:1}.}
        \label{fig:myREF_walkthrough}
        \end{figure*}

\subsection{Introduction to FLARE}  
    
The proposed FLARE algorithm develops sparsification-type solution with error accumulation to transmit low-dimensional representation of the model at each iteration. It uses a novel technique that manipulates the objective loss without using intensive communication or computational resources required to update the full high-dimensional model. As each client's accumulator is solely intended to store unsent weight updates during training, compensating for sparse communication by transmitting delayed updates once they accumulate sufficiently, each client locally retains accumulator data containing significant and unused optimization information. This information should be leveraged during the in-round optimization steps. We make further use of its values for each client by defining a new and client-specific loss term during each communication round. This term is designed to both minimize the client objective loss and to address the staleness effect. It is used to regularize the weights updates with respect to weights that were not transmitted, tilting the optimization track of each client closer to its original uncompressed track. 

\subsection{Algorithm Description}
\label{ssec:algorithm}

At the $k$th iteration round, each client transmits only the Top-$K$ updates (based on magnitude) and accumulates the error locally. At each round, we allow each client to conduct $p$ forward-backward passes according to $\tilde{f^i_k}$ in \eqref{eqn:6} below. We frequently denote the implementation of FLARE as $p$-FLARE to specify the parameter $p$ utilized in the current implementation of FLARE. Following the $p$ steps, each client transitions to the original and intended loss function denoted by $f$ for the remaining steps, employing any desired optimizer.
\begin{equation}
\begin{array}{l}
\displaystyle
\tilde{f^i_k}(x,\bar{w};\bar{A}^i_k,\bar{w_k}) \vspace{0.2cm}\\ \hspace{0.5cm}
\displaystyle= f(x,\bar{w})
\displaystyle
+ \tau_k \sum^{d}_{j=1} m(a_j) \cdot |w_j-(w_{k,j}+a_j)|.
\end{array}
    \label{eqn:6}
\end{equation}
\noindent
Here, $w_j, a_j, w_{k,j}$ are $j$-th entries of $\bar{w}$, $\bar{A}_k^i$, $\bar{w}_k$, respectively (where we omit complex indexing notations when the context is clear), summed over all model weights, $\tau_k$ is a regularization constant possibly varies with time, and
    \begin{equation}
        m(a) =   
        \begin{cases}
            1,     & \text{if $|a| > a_0$}, \\
            0,     & \text{otherwise}.
        \end{cases}  
    \label{masking}
    \end{equation}
    
Algorithm 1 describes the FLARE procedure with SGD. At the beginning of each round, clients determine $\tilde{f^i_k}$ according to their accumulators. Then, each client executes a forward-backward update according to $\tilde{f^i_k}$ for $p$ steps. Then, they revert to the original loss function $f$ for the remaining steps. Note that regularization in this description is based on the $L_1$ norm, but it can also be applied using the $L_2$ norm. Our theoretical results are developed specifically with respect to the $L_2$ norm. We next demonstrate the consecutive rounds by the algorithm step by step.

    \begin{algorithm}
    \begin{algorithmic} 
    \caption{The FLARE procedure with SGD}
    \State  Initialize weights at PS;  $\bar{w}_0$ 
    \State  Initialize objective loss;  $f$
    \State  Initialize for all $ 1 	\le i 	\le N$ clients: \newline \indent accumulator for client $k$; $\bar{A}^i_0 \leftarrow 0$
    \newline \indent Input data set for client $i$; $\rchi_i \in \rchi $
    \newline In parallel for all $ 1 \le i \le N$ clients:  
    
    \For {$k = 1,...,T$ number of rounds}
    \State PS to Clients Broadcast: $\bar{w}^i_{k-1} \leftarrow \bar{w}_{k-1} $
    \State  $\bar{w}^i_{k} \leftarrow ClientUpdate(\bar{w}^i_{k-1};\bar{A}^i_{k-1},\rchi_i)$
    \State $\bar{{\Delta}w}^i_{k} \leftarrow \bar{w}^i_{k} - \bar{w}_{k-1}$  
    \State $\bar{A}^i_{k} \leftarrow {\Delta}\bar{w}^i_{k} + \bar{A}^i_{k-1}$ 
    \State $mask \leftarrow 1$ at where $|\bar{A}^i_k|$ largest $R$
    \State $mask \leftarrow 0$ at where $|\bar{A}^i_k|$ not largest $R$
    \State $\bar{S}^i_k \leftarrow \bar{A}^i_{k} \odot mask$
    \State $\bar{A}^i_k \leftarrow \bar{A}^i_{k} \odot mask\neg$
    \State SparseBroadcast($\bar{S}^i_k$)
    \State ServerAggregation($\bar{S}^i_k + \bar{w}_{k-1}$)
    \EndFor
    \\\hrulefill
      \Procedure{$ClientUpdate$}{$\bar{w};\bar{A}^i_{k},\rchi_i$}
        \State define $\tilde{f^i_k}(x,\bar{w};\bar{A}^i_{k})$ and $\tilde{F}^i_k(\bar{w};\chi^i)$ using~\eqref{eqn:6},~\eqref{eqn:8} 
        \State $\mathcal{B} \leftarrow \rchi_i$  \algorithmiccomment{(split $\rchi_i$ into batches of size $B$)} 
        \State $StepsCounter \leftarrow 0$
        \For {$e = 1,...,E$}
            \For {batch $b \in \mathcal{B}$}
                \If{$StepsCounter < p$}
                    \State 
                    {
                    $\bar{w} \leftarrow \bar{w}-\gamma\nabla\tilde{F}^i_k(\bar{w};b)$
                    }
                \Else
                    \State
                    {
                    $\bar{w} \leftarrow \bar{w}-\gamma\nabla F^i_k(\bar{w};b)$
                    }
                \EndIf
                \State {$StepsCounter \leftarrow +1$}
            \EndFor
        \EndFor
        \State  \Return {$\bar{w}$}
      \EndProcedure
    \end{algorithmic}
    \label{algo:1-SFLARE}
    \end{algorithm}

To facilitate a clear and concise intuitive explanation of FLARE, let us consider a case of two clients, convex objective loss surface, and a two-dimensional model parameterized by two weights. For purposes of exposition, each client performs a single optimization step (i.e., $E=1$, $B=\infty$), and uses a sparsity level of $50\%$ (i.e., $R=50\%$) in the distributed SGD procedure with $a_0$ set to zero. This means that at every transmission from clients to the PS, the dimension with large update is sent to the PS, where the small one is kept at the clients. The loss surfaces for each client, $F_1(w;\chi_1), F_2(w;\chi_2)$, can be seen in Fig.~\ref{fig:myREF_walkthrough} colored by red and green, respectively, and the global objective is colored by black. We refer to the two-dimensional model weights that constitute $\bar{w} = (w_1,w_2)$ as horizontal and vertical. 


The description is provided below in four steps, accompanied by Figs. \ref{fig:myREF_walkthrough/Slide1}-\ref{fig:myREF_walkthrough/Slide4}.
    \begin{itemize}
        \item \textbf{Stage 1 (see Fig~\ref{fig:myREF_walkthrough/Slide1}):} The PS broadcasts global initialized model $\bar{w}_{0}$ (represented by the black dot in Fig~\ref{fig:myREF_walkthrough/Slide1}).
        \item \textbf{Stage 2 (see Fig~\ref{fig:myREF_walkthrough/Slide2}):} The model $\bar{w}_{0}$ sent by the PS is received at the two clients. Each client (say client $i$) computes:
    
        \begin{equation}
        \begin{array}{l}
        \displaystyle
        \tilde{f^i_0}(x,\bar{w};\bar{A}^i_0,\bar{w_0}) \vspace{0.2cm}\\ \hspace{0.5cm}
        \displaystyle= f(x,\bar{w})
        \displaystyle
        + \tau_0 \sum^{2}_{j=1} m(a_j) \cdot |w_j-(w_{k,j}+a_j)|.
        \end{array}
           \label{eqn:1}
        \end{equation}

Note that in this initialization step, the second term on the RHS of \eqref{eqn:1} can be ignored as $\bar{A}^i_0 = 0$ (and consequently $m(a)=0$) for all clients. As a result, each client performs the optimization steps with respect to the following expression:        
\begin{equation}
F^i(\bar{w};\chi^i) = \frac{1}{|\chi^i|} \sum_{x \in \chi^i} f(x,\bar{w}).
\end{equation}  

Then, each client transmits the low-dimensional embedding of $R$ portion of its most significant updates and locally accumulates the remaining weights. Illustrated in the two-dimensional example in Fig~\ref{fig:myREF_walkthrough/Slide2}, the horizontal updates are transmitted by both clients since they are the larger ones in this step, while the vertical updates are retained.
\item \textbf{Stage 3 (see Fig~\ref{fig:myREF_walkthrough/Slide3}):} The PS aggregates the individual updates from both clients by averaging the two lower solid-colored dots (horizontal) and derives an updated global model $\bar{w}_1$, as depicted in Fig~\ref{fig:myREF_walkthrough/Slide3}. Subsequently, the PS broadcasts $\bar{w}_1$ to the clients for the next iteration round. In the figure, the colored bars represent the accumulator information kept at the clients, with the upper colored dots signifying the complete local update before PS aggregation.
    \end{itemize}

Prior to explaining Stage 4, Let us look at the terms $\bar{w}_{1}+\bar{A}^i_1$, $i=1, 2$, marked in Fig~\ref{fig:myREF_walkthrough/Slide4} by $x$ dots. These terms represent the new models of the clients which can be regarded as if we only let $R$ portions of weights to be averaged at the PS aggregation, and keep the remaining weights unharmed by the sparsification. By minimizing the terms in \eqref{eqn:2} below, we further pull the next updates towards these new models, individually for each client:
\begin{equation}
     \tau_1||\bar{w}-(\bar{w}_{1}+\bar{A}^i_1)||^2_1 = \tau_1 \sum^{2}_{j=1} |w_j-(w_{1,j}+a_j)|
     \label{eqn:2}
    \end{equation} 
    
Minimizing \eqref{eqn:2} allows for the compensation of staled updates at each optimization step utilizing residual information, without any communication or computational costs. While the minimization of \eqref{eqn:2} influences both weights, we intended to pull only staled updates, i.e., the vertical update in our example at this stage. Pulling the horizontal weights is unwanted and may impede learning. As all of the horizontal accumulated content was just released, its accumulated value equals zero, leaving the horizontal weight to be regularized according to $|w-w_1|$ which serves no purpose but to slow convergence\footnote{The accumulator can be served as an indicator of how staled is an update at a specific round. The larger its accumulator content, the more staled it is.}. To tackle this issue, a masking operation is performed by detecting those up-to-date weights according to their accumulated values by the function $m(a)$, defined by \eqref{masking}. The expression $|a|>0$ indicates the weight is staled and should be pulled. With this, pulling is done only for staled updates. In addition, a new hyper parameter $\tau_k$ is set to control the pulling and the loss objective ratio.

After providing this explanation, we can now proceed to describe Stage 4:
        
    \begin{itemize} 
        \item \textbf{Stage 4 (see Fig~\ref{fig:myREF_walkthrough/Slide4}):} The updated global model $\bar{w}_1$  is received by the two clients. Then, each client (say client $i$) performs an optimization step with respect to the following loss:         
        \begin{equation}
        \begin{array}{l}
        \displaystyle
        \tilde{f^i_1}(x,\bar{w};\bar{A}^i_1,\bar{w_1}) \vspace{0.2cm}\\ \hspace{0.5cm}
        \displaystyle= f(x,\bar{w})
        \displaystyle
        + \tau_1 \sum^2_{j=1} m(a_j) \cdot |w_j-(w_{1,j}+a_j)|,
        \end{array}
           \label{eqn:5}
        \end{equation}
and,
\begin{equation}
        \tilde{F}^i_1(\bar{w};\chi^i) = \frac{1}{|\chi^i|} \sum_{x \in \chi^i} \tilde{f^i_1}(x,\bar{w};\bar{A}^i_1,\bar{w_1}).  
        \end{equation}
    \end{itemize}

This process repeats, where at each round each client performs an optimization step according to: 
    \begin{equation}
    \tilde{F}^i_k(\bar{w};\chi^i) = \frac{1}{|\chi^i|} \sum_{x \in \chi^i} \tilde{f^i_k}(x,\bar{w};\bar{A}^i_k,\bar{w_1}).
    \label{eqn:8}
    \end{equation}

\subsection{Further Insights and Observations of FLARE Algorithm}

\noindent
\textbf{Advantages of accumulated pulling and masking in FLARE:} The accumulated pulling implemented by FLARE helps each client to compensate for an unsent update at each round, and not only at the round the update is released, giving a direct solution to the staleness effect. Secondly, it creates a chain reaction:  More accurate accumulation in the second round aids each client in generating a new model with greater precision in the third round, enhancing the third accumulation, which, in turn, improves the fourth, and so forth. Lastly, masking can be done in various ways. To mitigate the regularization effect, masking can be defined more generally using a threshold $a_0$ as in \eqref{masking}. 
The larger $a_0$, the less weights are affected by FLARE. The threshold $a_0$ can be set for example as the average of all accumulator values.\vspace{0.2cm}


\noindent    
\textbf{The effect of $E$ values in FLARE:} In the previously presented example, each client performed only one forward-backward optimization step (i.e., $E=1$, $B$ is extremely large). When transitioning to larger $E$ values (or smaller $B$ values), FLARE executes more than one step for each client, and the following issue needs to be addressed. Firstly, by minimizing the same term in \eqref{eqn:5} for multiple forward-backward steps, there may be irrelevant and unnecessary weights influencing later steps. As the added regularization term remains unchanged during each round, weights may have already been compensated for being stale in the initial optimization steps. Minimizing \eqref{eqn:5} at each round emphasizes the significance of the initial steps in compensating for staleness. However, in the later steps, minimizing \eqref{eqn:5} might pull non-stale weights toward irrelevant points, resulting in incorrect updates. Consequently, each client's accumulator could become contaminated with less relevant residuals. These contaminated residuals accumulate, not only harming convergence by releasing incorrect updates once they reach a sufficient magnitude, but they also impact the regularization term itself at each round. This is because the regularization term depends on each accumulator's content, accelerating their loss of relevance.

To address this issue, we utilize the FLARE regularization term only for the first $p$ optimization steps in each round. In each round, only the initial $p$ steps are optimized according to $\tilde{f^i_k}(x,\bar{w};\bar{A}^i_k,\bar{w}^i_k)$, while the subsequent steps are optimized according to $f(x,\bar{w})$. This approach allows only the initial steps to correct for stale updates while preventing the rest from being influenced by any pulling that could lead to contamination. Additionally, we introduce a decay in the parameter $\tau$ as a function of $k$ with an exponential decay rule, setting $\tau_{k+1} = \tau_{k}/{c}$. These adjustments enhance the applicability of FLARE for multiple steps, mitigating the impact of contaminated accumulators on the regularization term. These observations are demonstrated in the experiments outlined in Section \ref{sec:experiments}.

\section{Convergence Analysis}
In this section we analyze the convergence of FLARE. Let $\delta$ be the sparsification ratio, $0<\delta<1$, and $C(\bar{w})$ be the Top-$K$ sparsification operator. Consistent with the assumptions used in \cite{pmlr-v97-karimireddy19a,NEURIPS2018_b440509a} and related studies, we introduce the following assumptions for our analysis: Assumption 1 pertains to the compressor, Assumption 2 addresses smoothness, and Assumption 3 concerns the bounded stochastic gradient moment.\vspace{-0.1cm}
\begin{assumption}
A Compression operator $C: \mathbb{R}^{d} \rightarrow  \mathbb{R}^{d}$ is a $\delta$ approximation for $0<\delta < 1$ if for all $\bar{w} \in \mathbb{R}^d$, the following inequality holds: $
||C(\bar{w})-\bar{w}||^2\le(1-\delta) ||\bar{w}||^2.$\vspace{-0.3cm}
\label{assumption1}
\end{assumption}
\begin{assumption}
Function \( f: \mathbb{R}^d \to \mathbb{R} \) is \( L \)-smooth if there exists a constant \( L > 0 \) such that for all \(\bar{x}, \bar{y} \in \mathbb{R}^d \), the following inequality holds:
$f(\bar{y}) \leq f(\bar{x}) + \nabla f(\bar{x})^T (\bar{y} - \bar{x}) + \frac{L}{2} \|\bar{y} - \bar{x}\|^2.$\vspace{-0.1cm}
\label{assumption2}
\end{assumption}
\begin{assumption} For any $\bar{w} \in \mathbb{R}^d$, the stochastic gradient fulfills: $\mathbb{E}||\nabla f(\bar{w})||^2 \le \sigma^2.$ 
\label{assumption3}
\end{assumption}
We begin by reviewing the analytic convergence performance achieved by the EC and sparsification method in \cite{pmlr-v97-karimireddy19a}. As detailed in \cite{NEURIPS2018_b440509a, pmlr-v97-karimireddy19a}, the error correction and sparsification methods are analytically validated for a single client. These methods, however, provide significant benefits when extended to multi-client settings, greatly enhancing their overall performance. It is also developed for a single optimization step at each iteration (i.e., FedSGD setting). In our paper, we adopt similar assumptions for the analysis of our proposed method. Specifically, we set $N=1, E=1, B=\infty$, $a_0=0$ in FLARE algorithm. Throughout the analysis, we refer to the loss function by $F$, and the regularized loss function under FLARE by $\tilde{F}$, where $\tilde{F}(\bar{w};\bar{w}_t,\bar{A}_t) = F(\bar{w}) + \frac{\tau}{2} || \Bar{w} - (\bar{w}_t + \bar{A}_t)||^2.$

Consider the case where $F$ is convex, and Assumptions \ref{assumption1} and \ref{assumption3} hold. Let $\bar{w}_0$ be the initial model, $\{\bar{w}_t\}_{t>0}$ be the sequence of updated model iterations produced by EC method \cite{pmlr-v97-karimireddy19a}, with step size $\gamma$, and let $\bar{w}^{avg}_T=\frac{1}{T}\sum^{T}_{t=0}{\bar{w}_t}$ be the averaged update. Then, the error under EC is bounded by \cite{pmlr-v97-karimireddy19a}:
\begin{align}
\hspace{-0.1cm}\mathbb{E}[F(\bar{w}^{avg}_T)] - F^{*} \le \dfrac{||\bar{w}_0 - w^{*}||^2}{2\gamma(T+1)} + \gamma \sigma^2 \left(\dfrac{1}{2} + \dfrac{2 \sqrt{1-\delta}}{\delta}\right),
\label{EF_convex_error}
\end{align}
where $w^{*}$ is the minimizer of $F$, and $F^*=F(w^{*})$.

\noindent
Secondly, in the non-convex scenario, assuming Assumptions \ref{assumption1}, \ref{assumption2}, and \ref{assumption3} hold, it can be shown that \cite{pmlr-v97-karimireddy19a}:
\begin{align}
\min_{t \in [T]} \mathbb{E}||\nabla F(\bar{w}_{t})||^2  \le 
&\dfrac{2(F(\bar{w}_0) - F(w^*))}{\gamma(T+1)}  +  L\gamma\sigma^2 +  \notag \\
& \hspace{10ex} \dfrac{4 \gamma^2 L^2 (1-\delta) \sigma^2}{\delta^2}.
\label{EF_non_convex}
\end{align}
Setting $\gamma$ on the order of $\frac{1}{\sqrt{T+1}}$ achieves the best convergence rate order $O(1/\sqrt{T})$ with $T$ as will be demonstrated later.

Setting $\delta=1$ results in the convergence rate of SGD\cite{NEURIPS2018_b440509a, pmlr-v97-karimireddy19a}. Furthermore, by choosing $\gamma$ on the order of $\frac{1}{\sqrt{T+1}}$ while keeping $\delta$ fixed, the method asymptotically approaches the performance of SGD as $T$ increases. However, in practical scenarios for finite $T$, the influence of $\delta$-related terms becomes significant. In particular, reducing
$\delta$ to increase the sparsification level can degrade the performance of EC through these terms. Specifically, when examining the scaling rate with respect to the sparsification ratio $\delta$ while keeping $T$ finite, it becomes evident that the error bound in \eqref{EF_convex_error} scales with a rate of $1/\delta$, and the gradient moment in \eqref{EF_non_convex} scales with a rate of $1/\delta^2$, as $\delta$ approaches zero. This degradation is notably pronounced under extreme sparsification settings, where the performance of EC significantly deteriorates. 

In the following theorems, we demonstrate that FLARE offers substantial performance enhancements, particularly in mitigating the performance degradation associated with increased sparsification levels. Specifically, we show below that for the error bound in the convex case scales with a rate of $1/\sqrt{\delta}$, and the gradient moment in the non-convex case scales with a rate of $1/\delta$, as $\delta$ approaches zero. Furthermore, FLARE achieves the same order of $1/\sqrt{T}$ with $T$ as EC.
    \begin{figure}
    \centering
    \includegraphics[width=0.3\textwidth]{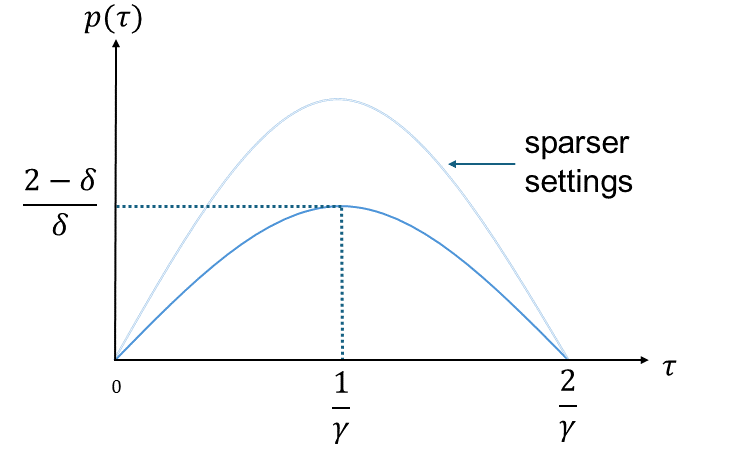}
    \caption{An illustration of the function $p(\tau)$.}
    \label{p_mu}
    \end{figure}

\begin{theorem}
Consider the case where Assumption \ref{assumption1}, \ref{assumption3} hold, and the loss function $F$ is convex. Denote the model iterates of FLARE by $\{ \bar{w}_t\}_{t>0}$, with step size $\gamma$, and let $\bar{w}^{avg}_T=\frac{1}{T}\sum^{T}_{t=0}{\bar{w}_t}$. Denote the minimizer of $\tilde{F}$ by $w^{*}$,and let $\tilde{F}^*=\tilde{F}(w^{*})$. Then, 
\begin{align}
&\mathbb{E}[\tilde{F}(\bar{w}^{avg}_T)] - \tilde{F}^{*} \le \dfrac{||\bar{w}_0 - w^{*}||^2}{2\gamma(T+1)} +\notag \\
&\hspace{20ex} \gamma \sigma^2 \left(\dfrac{1}{2} + \dfrac{2 \sqrt{1-\delta}}{\delta \sqrt{1+p(\tau)}}\right),
\label{Theorem1:convex_objective}
\end{align}
where 
\begin{align}
p(\tau) = \dfrac{2 \gamma }{\delta} (2-\delta) \tau + \dfrac{2 \gamma^2 }{\delta} (\dfrac{\delta}{2}-1) \tau^2.
\label{p_tau}
\end{align}
\label{Theorem1}
\end{theorem}
\noindent
\noindent
The proof can be found in the Appendix.

Next, we set $\gamma = \dfrac{a}{\sqrt{T+1}}$ as done in \cite{pmlr-v97-karimireddy19a}, with $a=\nicefrac{||\bar{w}_0-w^{*}||}{\sigma}$, to achieve the best convergence rate with $T$. Then, the error bound under FLARE \eqref{Theorem1:convex_objective} becomes: 
\begin{align}
 \mathbb{E}[\tilde{F}(\bar{w}^{avg}_T)] - \tilde{F}^{*} \le
&\frac{\left\|\bar{w}_0 - w^*\right\|}{2 \sqrt{T+1}} \sigma \left(2+\frac{4 \sqrt{1-\delta}}{ \sqrt{\delta^2(1+p(\tau))}}\right).
\end{align}
\label{remark1: convex_with_gamma}
For comparison, the error bound under EC \eqref{EF_convex_error} becomes:
\begin{align}
\mathbb{E}[F(\bar{w}^{avg}_T)] - F^{*} \le \dfrac{||\bar{w}_0 - w^{*}||}{2\sqrt{T+1}} \sigma  \left( 2 + \dfrac{4\sqrt{1-\delta}}{\sqrt{\delta^2}  }  \right).
\label{EF rate}
\end{align}

As illustrated in Fig.~\ref{p_mu}, the function $p(\tau)$ is parabolic with $p(\tau) > 0$ for $0< \tau < \frac{2}{\gamma}$ with maximum at $\nicefrac{(2-\delta)}{\delta}$. For $\tau=0$, $p(0) = 0$, which aligns with the established convergence rate of the EC method as shown in \eqref{EF_convex_error}. However, our approach improves upon the EC method by leveraging the FLARE regularization coefficient to reduce the error scaling. Specifically, the positive factor $\sqrt{1+p(\tau)}$ helps to mitigate the scaling with $\delta$. By choosing $\tau$ such that $p(\tau)$ scales as $1/\delta$, FLARE achieves an error bound that scales as $1/\sqrt{\delta}$, in contrast to the $1/\delta$ scaling under EC. Note that as $\delta$ approaches zero, for a finite $T$, $\tau$ can be arbitrarily small, by setting: $\tau=1/(\gamma\cdot C)$ for any fixed $C>1$, yielding $p(\tau)=\frac{2C-1}{C^2}\cdot\frac{2-\delta}{\delta}$ scales with $1/\delta$. Additionally, in terms of the scaling order with $T$, FLARE maintains the same order of $1/\sqrt{T}$ as achieved by EC. 

\begin{theorem}
Consider the case where Assumption \ref{assumption1},\ref{assumption2} and \ref{assumption3} hold. Denote the model iterates of FLARE by $\{ \bar{w}_t\}_{t>0}$, with step size $\gamma$. Denote the minimizer of $\tilde{F}$ by $w^{*}$. Then,
\begin{align}
&\min_{t \in [T]} \mathbb{E} ||\nabla \tilde{F}(\bar{w}_{t+1})||^2  \le  \dfrac{2(F(\bar{w}_0) - F(w^{*}))}{\gamma(T+1)} +  L\gamma\sigma^2 + \notag \\
& \hspace{5ex} \dfrac{2}{\gamma}  \left( \dfrac{L^2 \gamma}{2} + r(\tau)\right) \dfrac{4\gamma^2 (1-\delta) \sigma^2}{\delta^2}  \cdot\dfrac{1}{1+p(\tau)},
\label{eq:th_gradient} 
\end{align}
where 
\begin{align}
&r(\tau) = \left( \dfrac{\gamma}{2} + \dfrac{L\gamma^2 }{2}\right) \tau^2.
\end{align}
\label{Theorem2}
\end{theorem}
The proof can be found in the Appendix.

\noindent
Note that $r(\tau) > 0$ for $\tau>0$. Again, we are aligned with the known rate of EC \eqref{EF_non_convex} for $\tau=0$.

Next, we set $\gamma = \dfrac{1}{\sqrt{T+1}}$ as done in \cite{pmlr-v97-karimireddy19a} to achieve the best convergence rate with $T$. Then, the gradient moment bound under FLARE \eqref{eq:th_gradient} becomes:
\begin{align}
&\min_{t \in [T]} \mathbb{E} ||\nabla \tilde{F}(\bar{w}_{t+1})||^2   \le  \dfrac{2(F(w_0) - F(w^{*}))+L\sigma^2}{\sqrt{T+1}} + \notag \\
&\dfrac{4L^2 (1-\delta) \sigma^2}{\delta^2 (T+1)} \cdot \dfrac{1}{1+p(\tau)}  +\dfrac{4 (1-\delta) \sigma^2}{(\nicefrac{\delta}{\tau})^2 (T+1)} \cdot \dfrac{1}{1+p(\tau)} \notag\\
&  \hspace{1.5cm}+\dfrac{2}{(\nicefrac{\delta}{\tau})^2} \dfrac{4 (1-\delta) \sigma^2}{\sqrt{T+1} (T+1)} \cdot \dfrac{1}{1+p(\tau)}.
\end{align}
For comparison, the gradient moment bound under EC \eqref{EF_non_convex} becomes:
\begin{align}
\min_{t \in [T]} \mathbb{E} ||\nabla F(&\bar{w}_{t+1})||^2\le   \dfrac{2(F(\bar{w}_0) - F(w^{*}))+L\sigma^2}{\sqrt{T+1}} \notag \\
& \hspace{2cm}+\dfrac{4 L^2 (1-\delta) \sigma^2}{\delta^2 (T+1)}.
\end{align}

For $\tau=0$, $p(0) = 0$, and the results are aligned with the known rate of EC \eqref{EF_convex_error} in this case as well. However, it can be seen again that our approach improves upon the EC method by leveraging the FLARE regularization coefficient to reduce the error scaling. By choosing $\tau$ such that $p(\tau)$ scales as $1/\delta$, FLARE achieves an error bound that scales as $1/\delta$, in contrast to the $1/\delta^2$ scaling under EC. Additionally, in terms of the scaling order with $T$, FLARE maintains the same order of $1/\sqrt{T}$ as achieved by EC.
\section{Experiments}
\label{sec:experiments}

    \begin{figure}[t]
         \centering
         \begin{subfigure}{0.35\textwidth}
            \includegraphics[width=\textwidth]{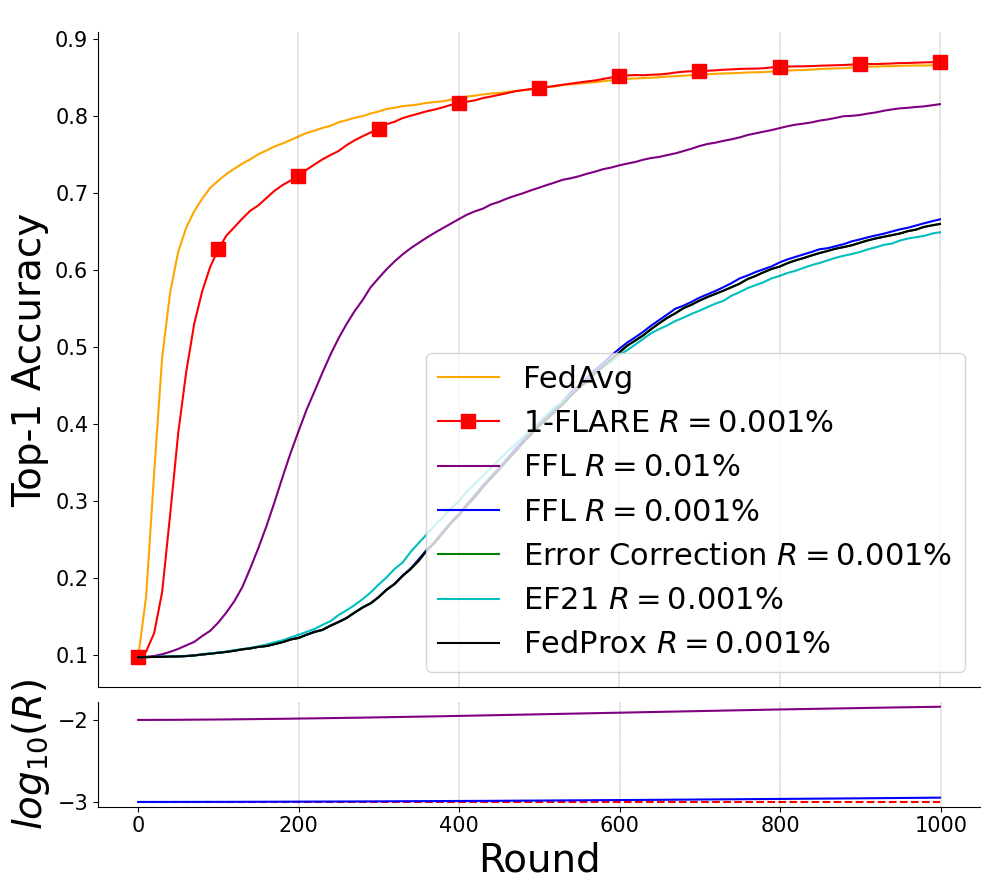}
            \caption{FC}
         \end{subfigure}
         \hfill
         \begin{subfigure}{0.35\textwidth}
            \includegraphics[width=\textwidth]{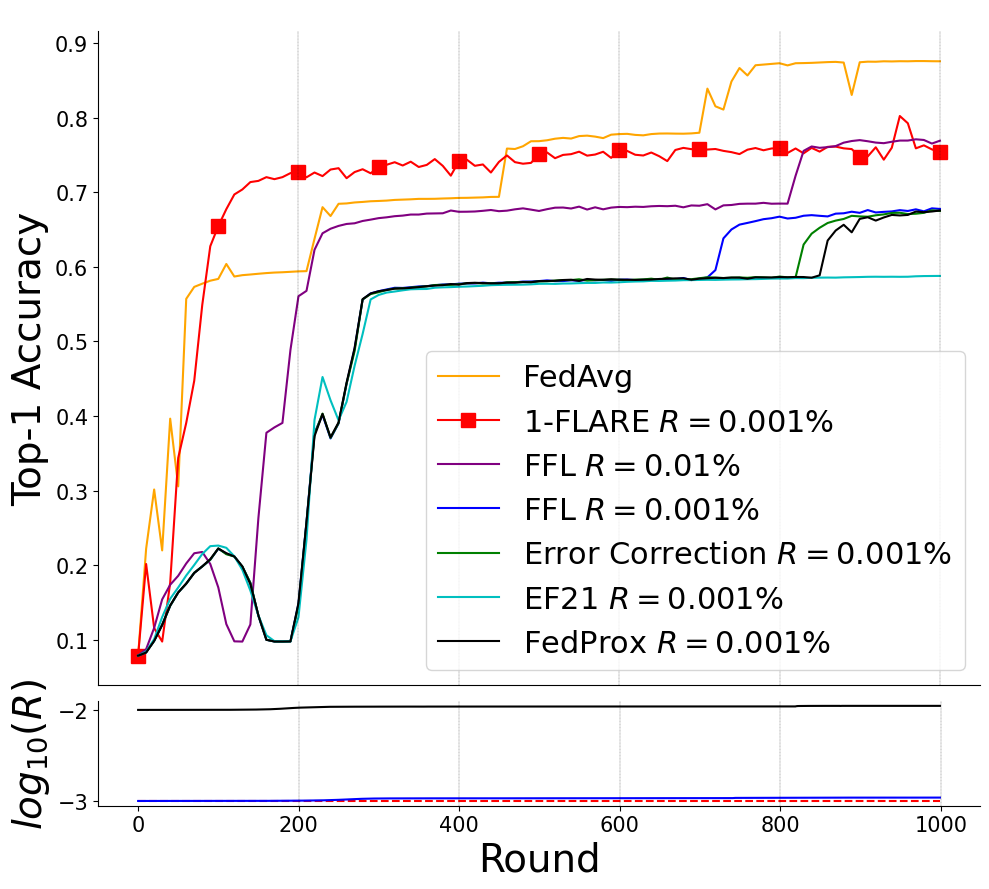}
            \caption{CNN}
         \end{subfigure}          
    \caption{The test accuracy performance is compared for FC and CNN models on MNIST with $E=1$ and $B=\infty$. $1$-FLARE is implemented with a sparsity setting of $R=0.001\%$. FFL, EF21, FedProx and Error Correction are included in the comparison. The uncompressed FedAvg is presented as a benchmark for performance. Remarkably, $1$-FLARE outperforms all other methods even when FFL is configured with a sparsity of $R=0.01\%$.}
    \label{1E}
    \end{figure}

    \begin{figure*}[t]
         \centering
         \begin{subfigure}{0.245\textwidth}
            \includegraphics[width=\textwidth]{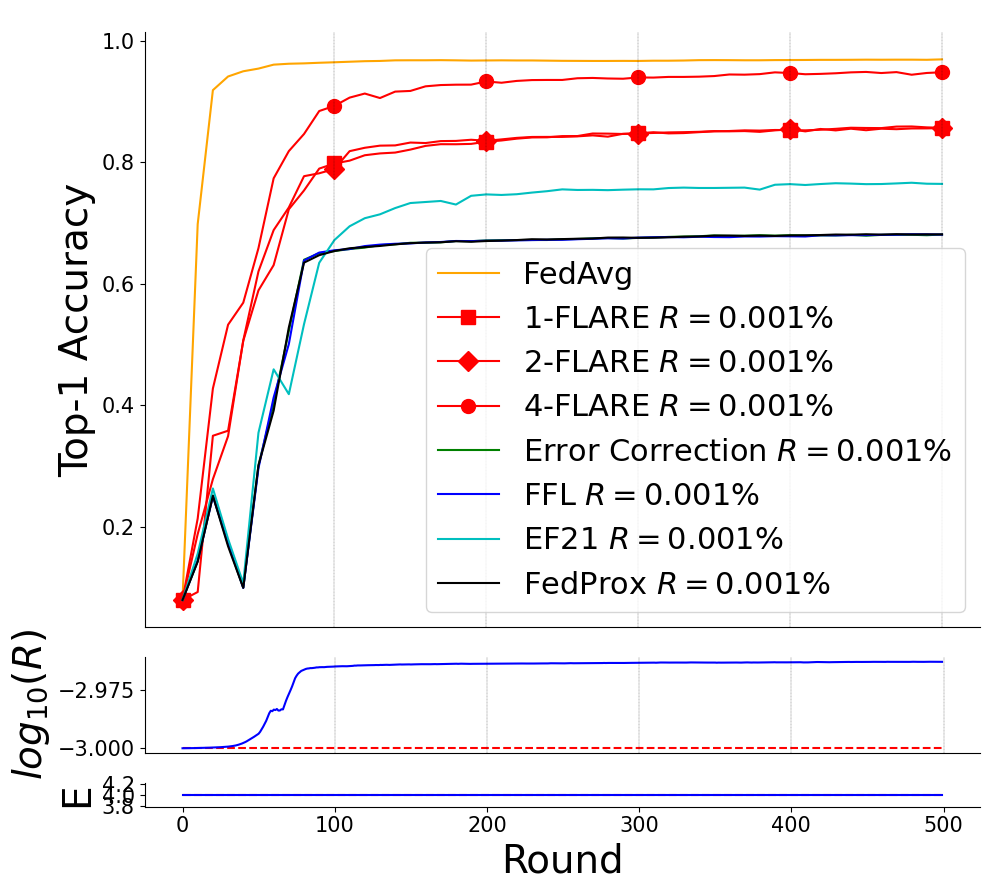}
            \caption{$E=4$}
         \end{subfigure}
         \begin{subfigure}{0.245\textwidth}
            \includegraphics[width=\textwidth]{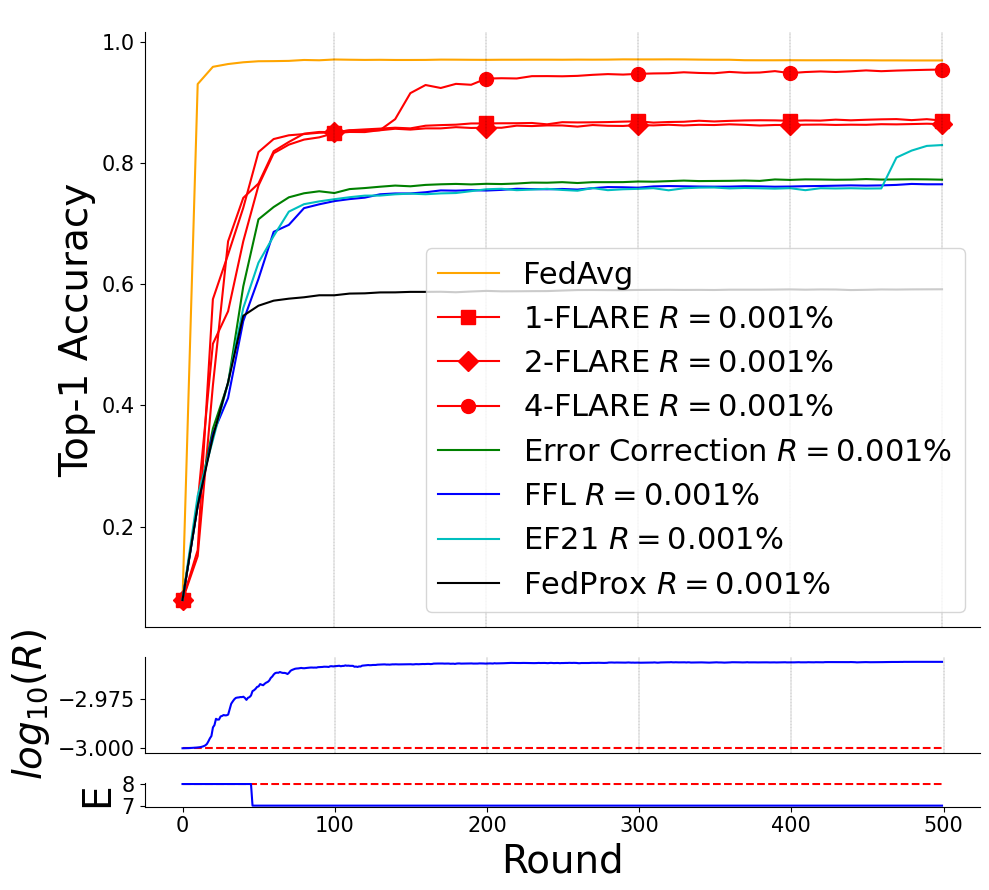}
            \caption{$E=8$}
         \end{subfigure}          
         \centering
         \begin{subfigure}{0.245\textwidth}
            \includegraphics[width=\textwidth]{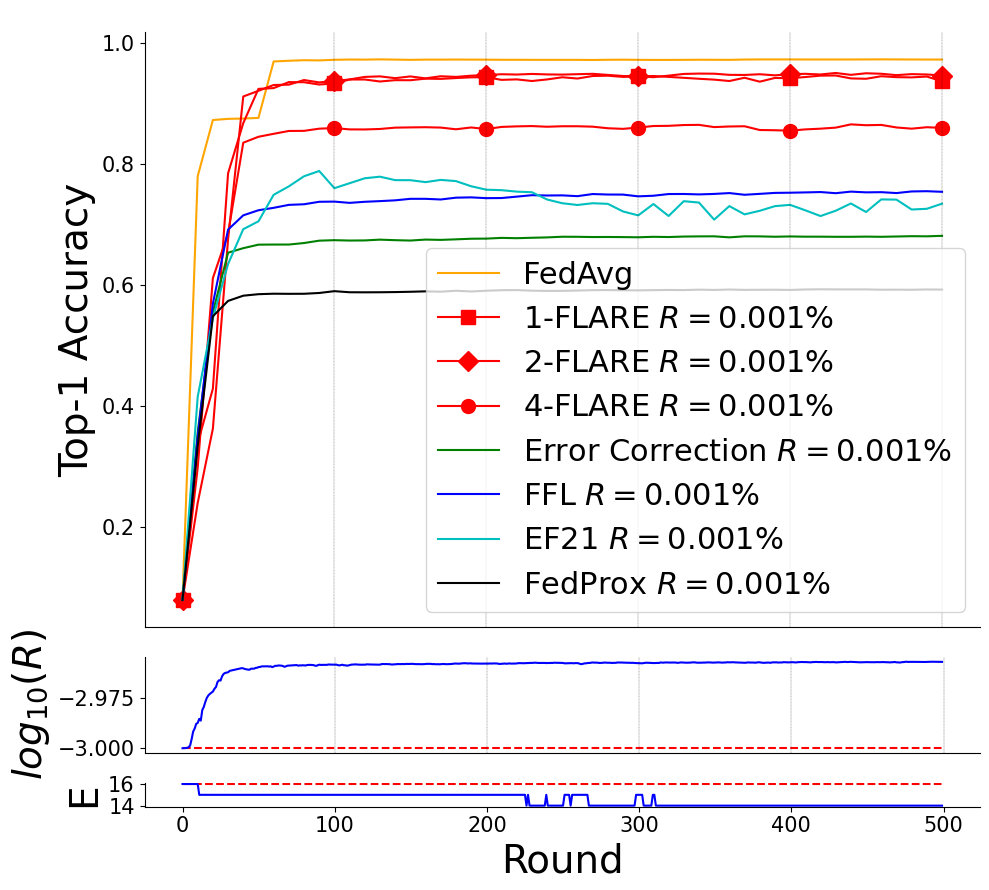}
            \caption{$E=16$}
         \end{subfigure}
         \begin{subfigure}{0.245\textwidth}
            \includegraphics[width=\textwidth]{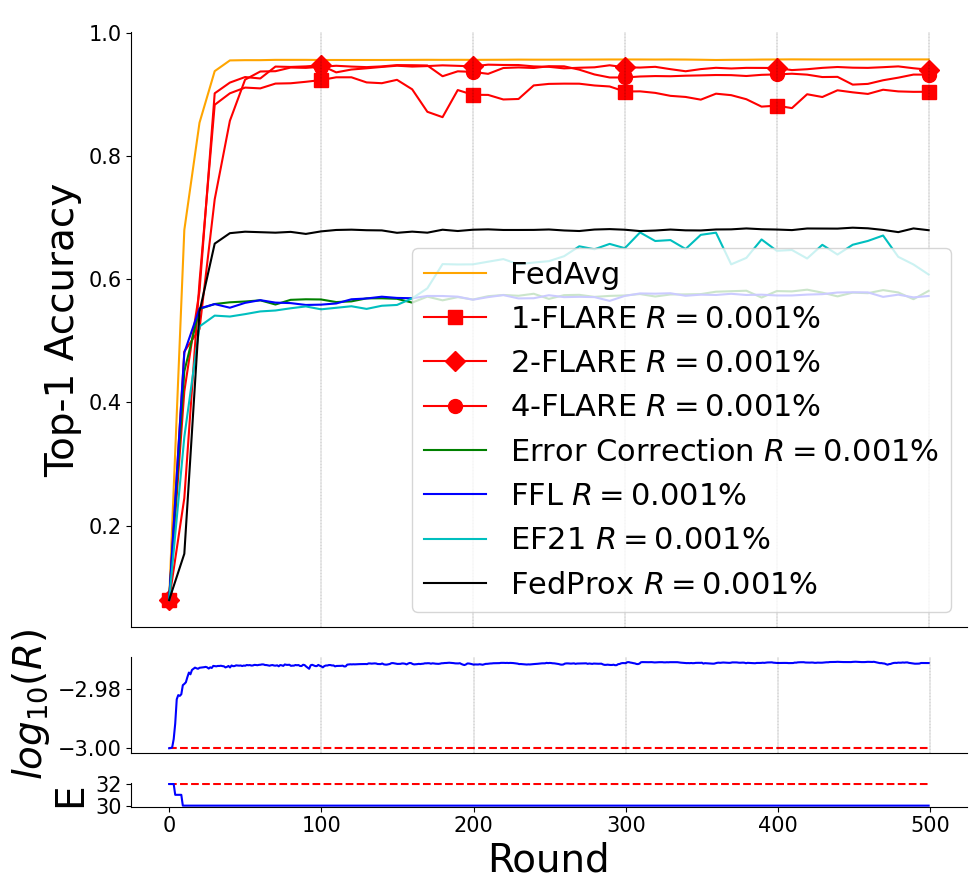}
            \caption{$E=32$}
         \end{subfigure}          
    \caption{The test accuracy performance is compared for CNN model on digit MNIST with $R=0.001\%$, $p=1,2,4$, $\tau=0.05$, $c=1.1$, 10 clients with $B=\infty$ and $E=4,8,16,32$. $p$-FLARE is compared with uncompressed FedAvg (benchmark for performance), Error Correction, FFL, EF21 and FedProx methods. At the bottom of each figure, $E$ and $log_{10}(R)$ values are plotted to illustrate the sparsity and computation levels of $p$-FLARE compared to FFL. It is evident that $p$-FLARE significantly outperforms all other methods in all cases.}
    \label{CNN}
    \end{figure*}

    \begin{figure*}[t]
         \centering
         \begin{subfigure}{0.245\textwidth}
            \includegraphics[width=\textwidth]{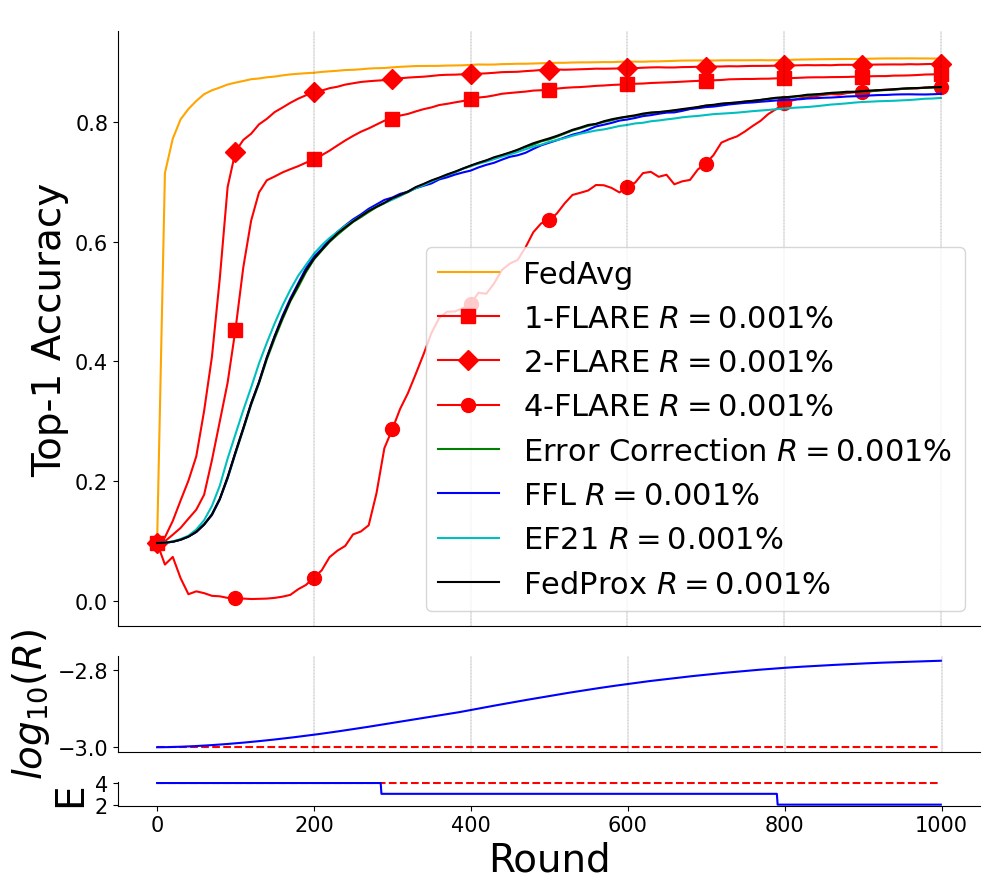}
            \caption{$E=4$}
         \end{subfigure}
         \begin{subfigure}{0.245\textwidth}
            \includegraphics[width=\textwidth]{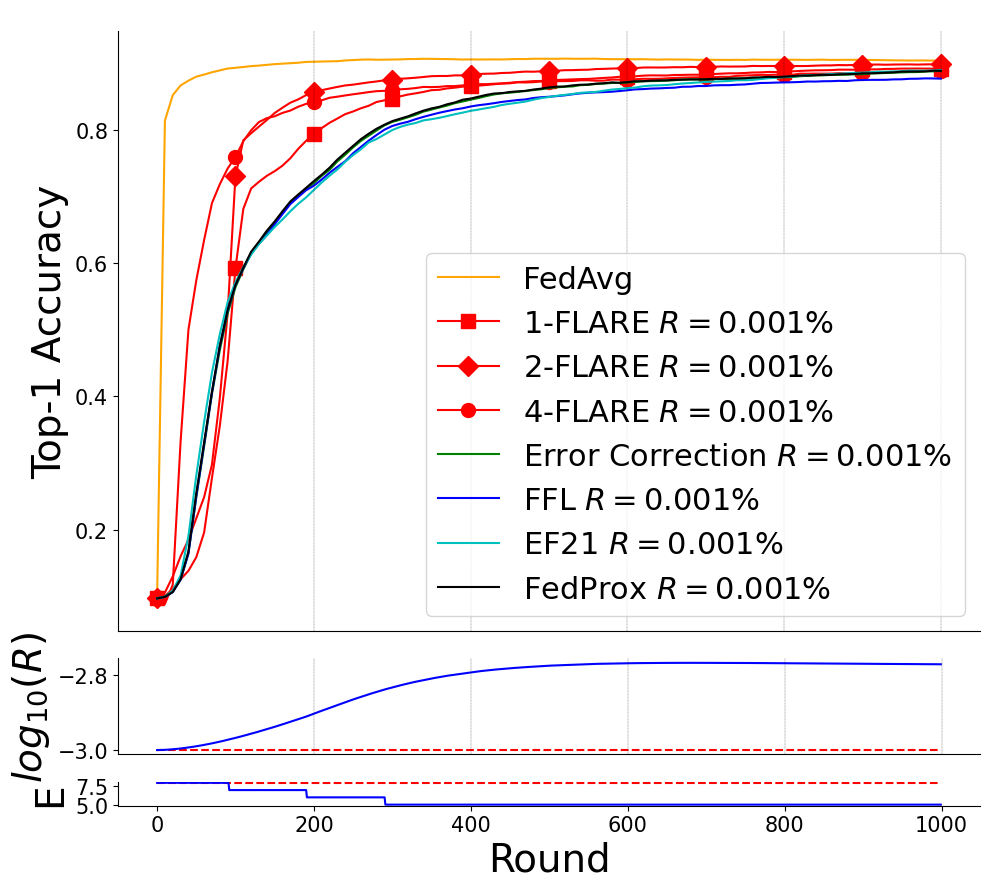}
            \caption{$E=8$}
         \end{subfigure}          
         \centering
         \begin{subfigure}{0.245\textwidth}
            \includegraphics[width=\textwidth]{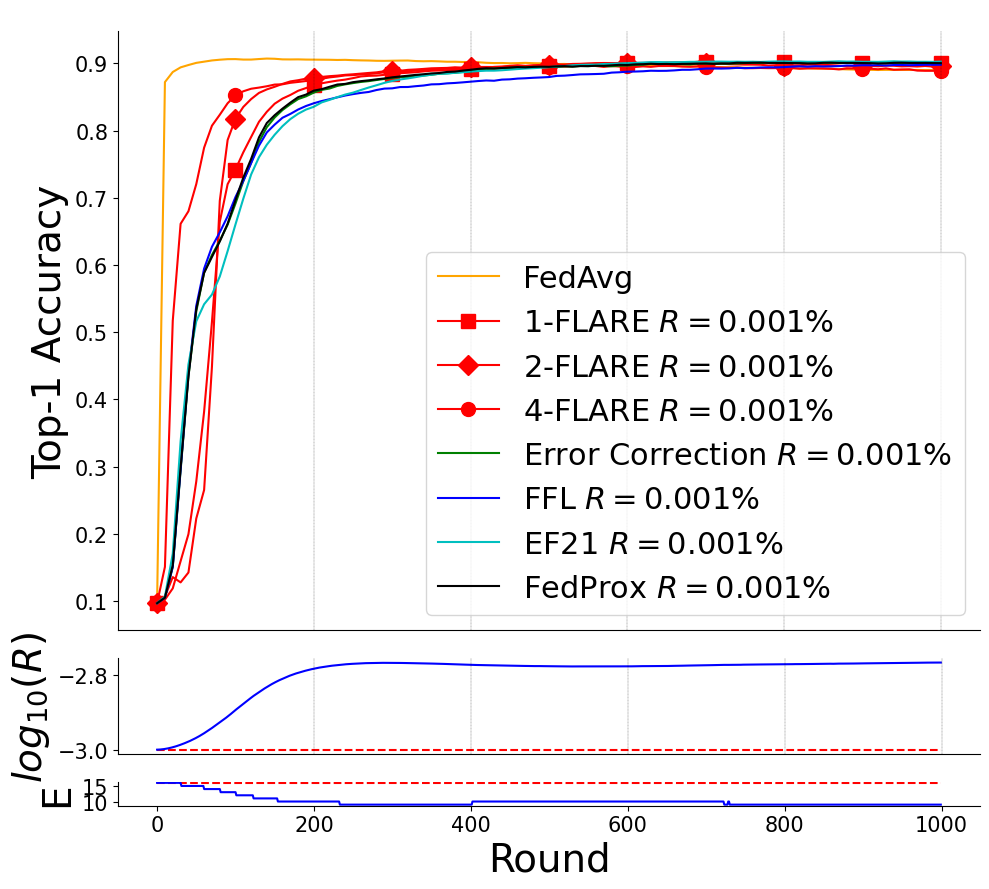}
            \caption{$E=16$}
         \end{subfigure}
         \begin{subfigure}{0.245\textwidth}
            \includegraphics[width=\textwidth]{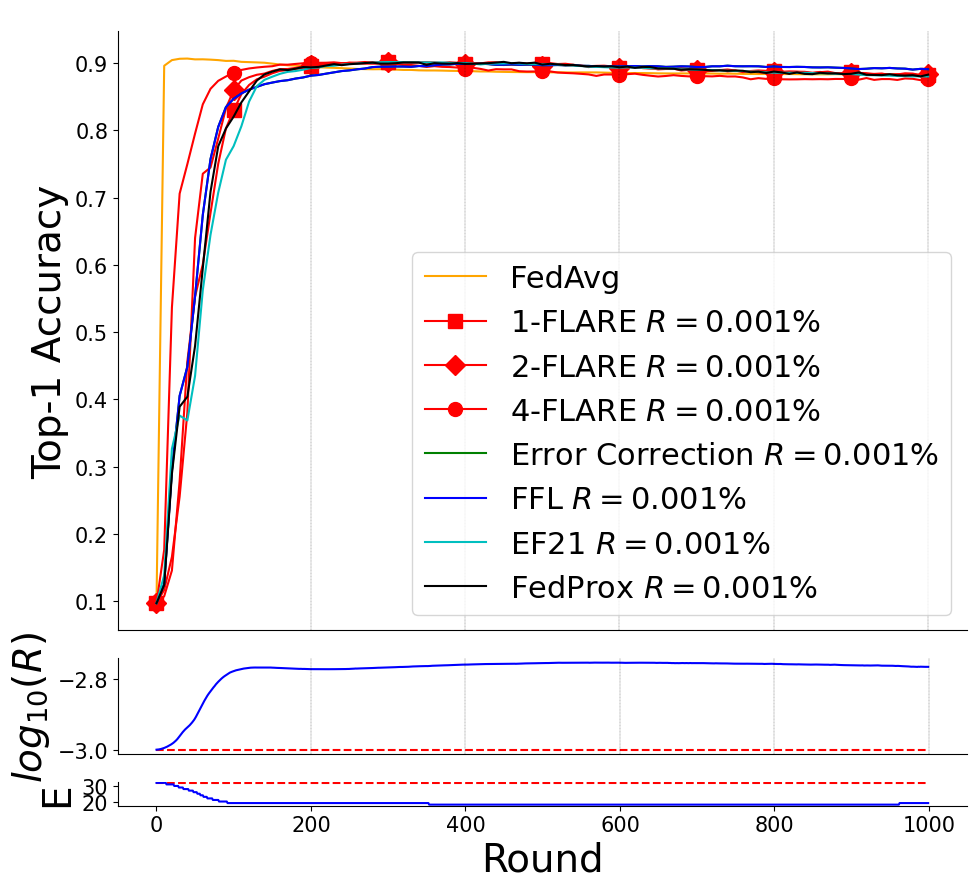}
            \caption{$E=32$}
         \end{subfigure}          
    \caption{The test accuracy performance is compared for FC model on digit MNIST with $R=0.001\%$, $p=1,2,4$, $\tau=0.5$, $c=1.05$, 10 clients with $B=\infty$ and $E=4,8,16,32$. $p$-FLARE is compared with uncompressed FedAvg (benchmark for performance), Error Correction, FFL, EF21 and FedProx methods. At the bottom of each figure, $E$ and $log_{10}(R)$ values are plotted to illustrate the sparsity and computation levels of $p$-FLARE compared to FFL. It is evident that $p$-FLARE outperforms all other methods in all cases in terms of convergence rate, demonstrating strong performance in accuracy as well.}
    \label{FC}
    \end{figure*}

    \begin{figure}[t]
         \centering
         \begin{subfigure}{0.45\textwidth}
            \includegraphics[width=\textwidth]{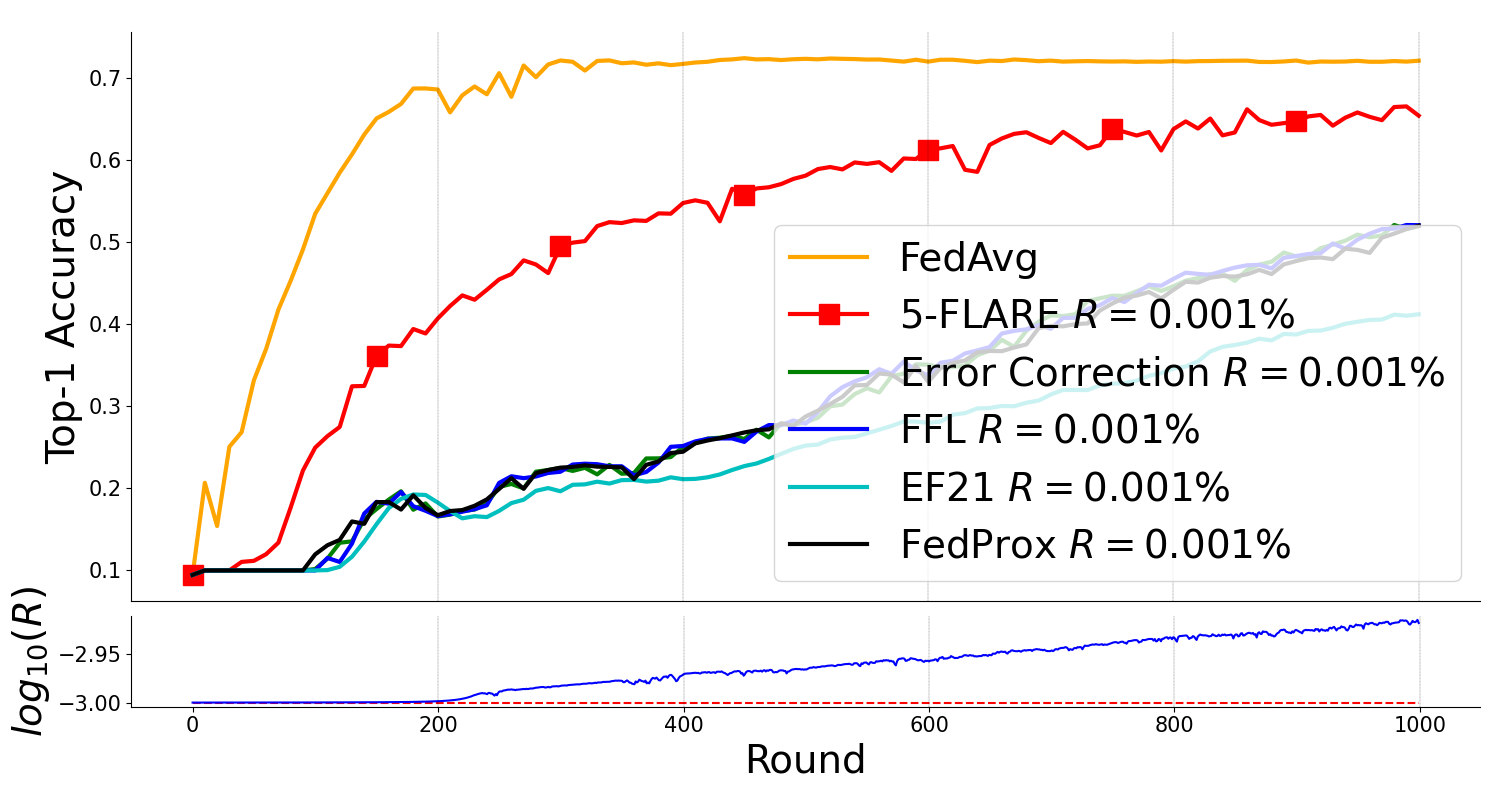}
            \caption{VGG11}
            \label{VGG11}
         \end{subfigure}
         \hfill
         \begin{subfigure}{0.45\textwidth}
            \includegraphics[width=\textwidth]{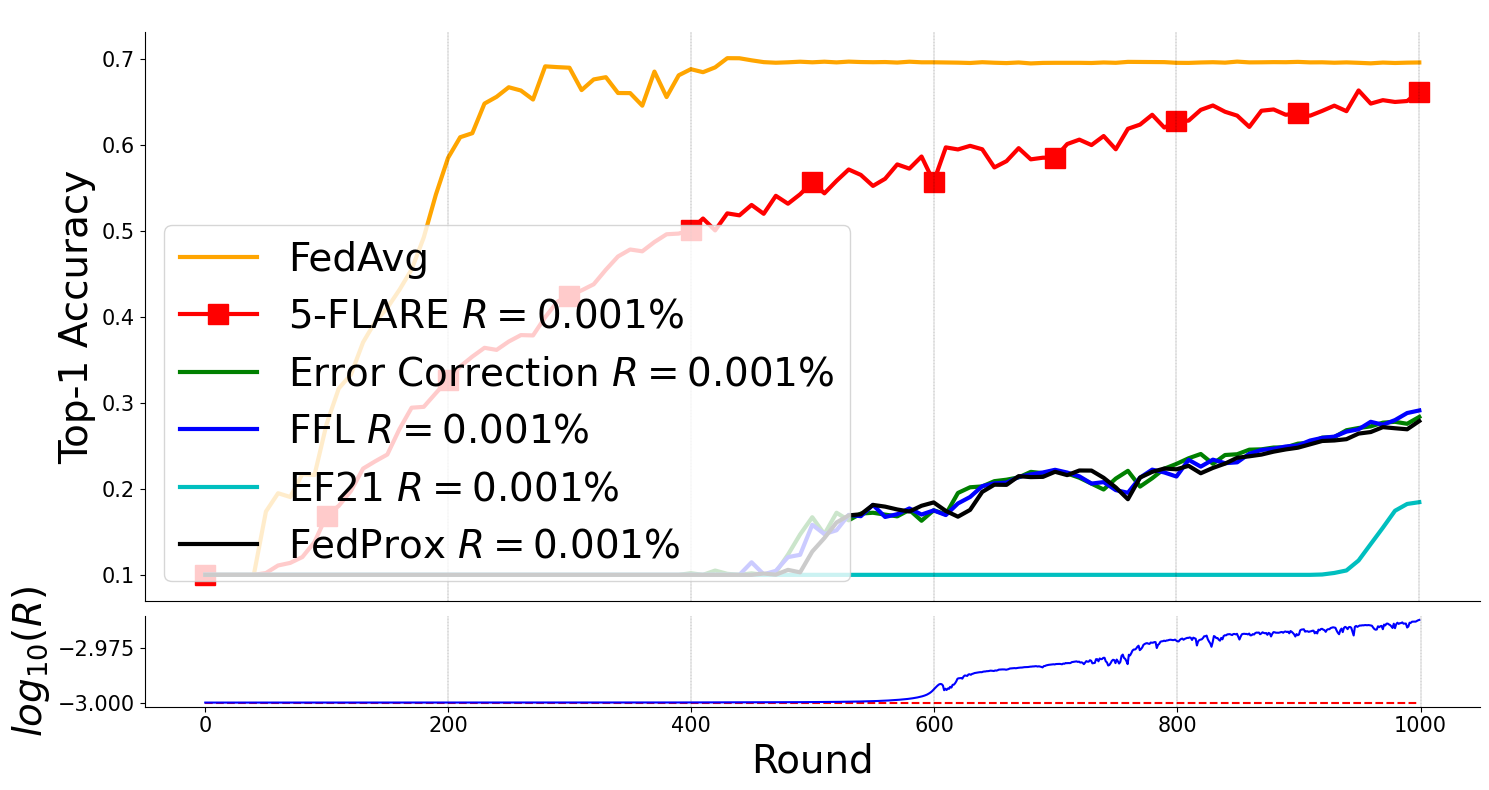}
            \caption{VGG16}
            \label{VGG16}
         \end{subfigure} 
         \begin{subfigure}{0.45\textwidth}
            \includegraphics[width=\textwidth]{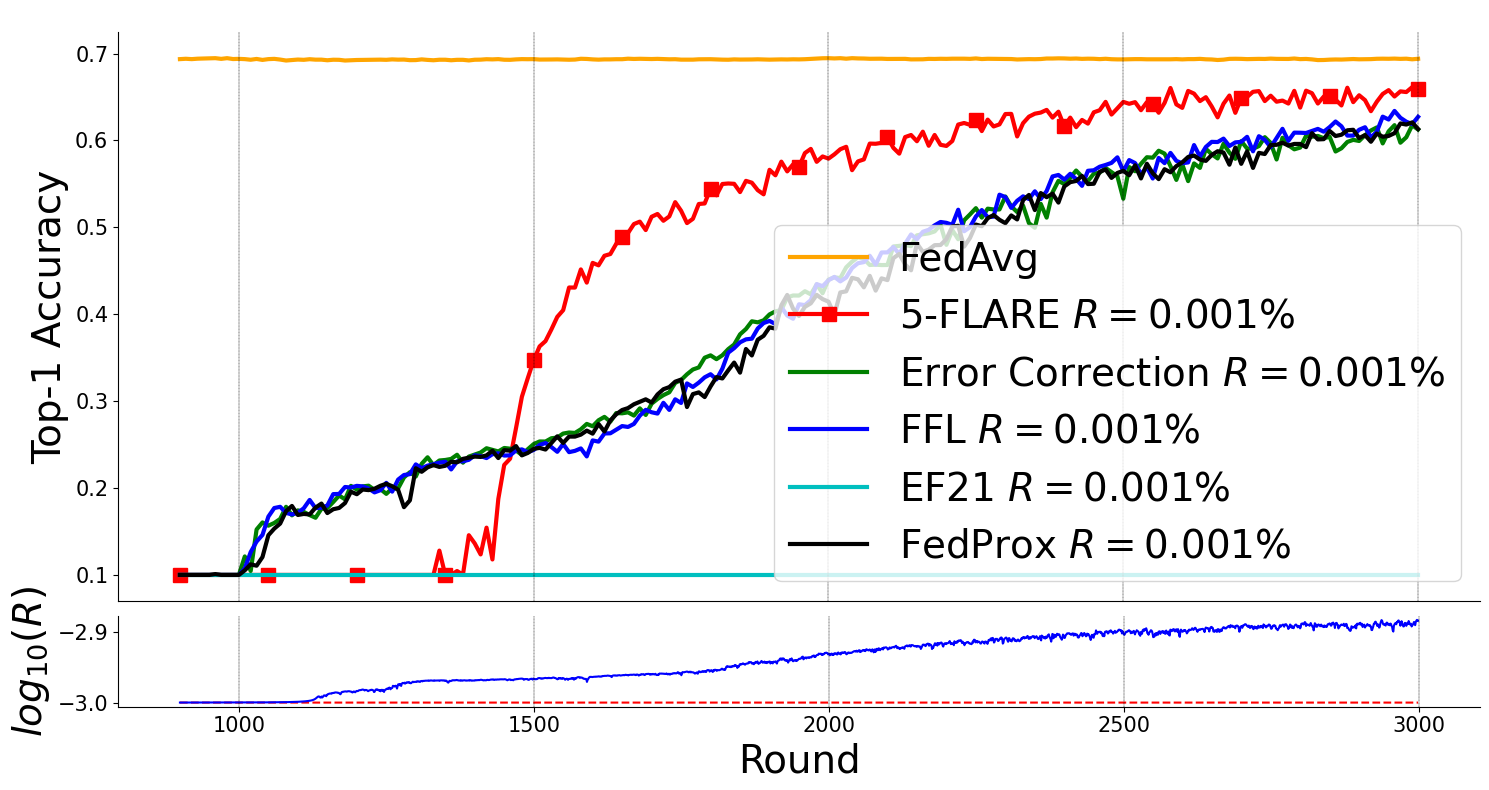}
            \caption{VGG19}
            \label{VGG19}
         \end{subfigure}           
    \caption{The test accuracy performance is compared for VGG models on CIFAR10 with $R=0.001\%$, $p=5$, $\tau=0.05$, $c=1.01$, 4 clients with 12500 examples and $B=500$ and $E=1$. $p$-FLARE is compared with uncompressed FedAvg (benchmark for performance), Error Correction, EF21, FedProx, and FFL methods. It is evident that $p$-FLARE significantly outperforms all other methods in all cases.}
    \label{VGG}
    \end{figure}

In this section, we present a series of simulations conducted to evaluate the performance of the proposed FLARE algorithm. We simulated an FL system with $N$ clients and a PS. The datasets were distributed across the clients to simulate the scenario of distributed nodes storing local data. The central unit performs computations at the PS. 

We conducted simulations for image classification and text generation tasks employing six distinct models across different FL settings: FC, CNN, VGG 11, VGG 16, VGG 19, and GRU models. Our initial evaluation focused on assessing the performance of FLARE using the FC and CNN models, trained on the MNIST digit dataset \cite{lecun1998gradient}. We extensively tested its performance under various settings to explore its robustness for both balanced and imbalanced data distributions. Subsequently, we scaled up to larger models, including VGG11, VGG16 (138.4M parameters), and VGG19 (143.7M parameters) \cite{simonyan2014very} trained on the CIFAR10 dataset \cite{krizhevsky2009learning}, and conducted a text generation task based on "The Complete Works of William Shakespeare" \cite{caldas2019leaf}. In all experiments, the models were initialized with the same global model, consistent across experiments, and evaluated on the test set every 10 rounds. 

Throughout the experiments, FLARE was configured with a sparsity setting up to $R=0.001\%$. We also conducted additional experiments to evaluate the performance with less aggressive sparsification settings. We compared our method with uncompressed FedAvg, the classic Error Correction method, and the state-of-the-art FFL, EF21 and FedProx algorithms \cite{DBLP:journals/corr/abs-1712-01887,NEURIPS2021_231141b3,MLSYS2020_1f5fe839}. (i) FFL is an adaptive method that adjusts two variables, $E$ and $R$, to efficiently conduct compression schemes such as Error Correction and further minimize the learning error. It has demonstrated superiority over other recent methods \cite{9809924}, including Qsparse, ATOMO, or AdaComm \cite{basu2019qsparse,wang2018atomo,wang2019adaptive}. Since FFL adjusts $E$ and $R$ during training, we include plots of $log_{10}(R)$ and $E$ (if it changes) at the bottom of each figure to verify that our method maintains greater sparsity throughout all rounds. (ii) EF21 ~\cite{NEURIPS2021_231141b3} is a newly enhanced version of the standard Error Feedback mechanism. It uses a recursive compressor to reduce compression errors throughout the training process. EF21 has demonstrated superior performance over other methods across a wide range of experiments, highlighting its effectiveness in leveraging error feedback to reduce communication costs. (iii) FedProx ~\cite{MLSYS2020_1f5fe839} is an FL algorithm that employs dedicated regularization during local training, and therefore a good benchmark for comparison. 
For our experiments, we integrated Top-K sparsification and error feedback with the regularization techniques of FedProx, ensuring that both FLARE and FedProx share the same objective.

\subsection{Experiment 1: FC and CNN Models on the MNIST Dataset}

The FC model consists of three hidden layers, each with 4069 neurons (identical setting as in\cite{aji-heafield-2017-sparse}). The CNN consists of two $5\times 5$ convolution layers, the first with 32 channels, the second with 64, each followed with $2\times 2$ max pooling (identical setting as in\cite{mcmahan2017communication}). For the FC and CNN experiments, the FL setting consists of 10 clients each holds 600 examples, global test set of 10000, and we use SGD with full batch size ($B$ is extremely large) for consistency. In order to assess the robustness of $p$-FLARE (where $p$ denotes the specific parameter used in FLARE), we conduct tests for $E=1,4,8,16,32$ separately, each with $p=1,2,4$. The results are presented in the same figure, as seen in Figs. \ref{1E}, \ref{CNN}, \ref{FC}. The $p$-FLARE for the FC model is set with $\tau=0.5$ and decay constant $c=1.05$, and the CNN experiments are set with $\tau=0.05$ and $c=1.1$. Throughout all experiments, we set $a_0$ to be the median of the accumulated error at absolute values, implying that pulling is applied to only $50\%$ of all model weights, considered to be the most staled based on their accumulator value.

Fig~\ref{1E} shows $1$-FLARE for CNN and FC models for the $E=1$ case. The FC model with $1$-FLARE shows to achieve sparsity of $99.999\%$ ($R=0.001\%$) converging with only a negligible delay without any Top-1 accuracy damage. The CNN model with $1$-FLARE shows to follow the curve of the uncompressed FedAvg without any delay, having small accuracy damage. $1$-FLARE outperforms FFL, EF21, FedProx and Error Correction, even when we set FFL with $R=0.01\%$ for both models. On the CNN model, FedAvg achieves Top-1 test accuracy of $0.78$, FLARE achieves $0.73$ while FFL with $R=0.01\%$ achieves $0.67$, meaning our method with one order sparser level achieves greater accuracy performances. On the FC model, FLARE accuracy curve merges with the FedAvg after 400 rounds while FFL with $R=0.01\%$ did not reach this point after 1000 rounds. 

Fig~\ref{CNN} and Fig~\ref{FC} show CNN and FC models results for $E=4,8,16,32$, where we use $p=1,2,4$. In Fig~\ref{CNN}, it is evident that CNN training is significantly impacted by sparsification with $R=0.001\%$ using any of the compared methods, while $p$-FLARE enables training with a reasonable delay. For example, on $E=32$, $4$-FLARE achieves $0.92$ Top-1 test accuracy, while Error Correction achieves only $0.58$.


\subsection{Experiment 2: VGG 11, 16, 19 Models on the CIFAR10 Dataset}

In this experiment, we validate $p$-FLARE on VGG 11, 16, and 19 models trained on the CIFAR-10 dataset with a sparsity setting of $R=0.001\%$. CIFAR-10 comprises 10 classes of $32\times 32$ images with three RGB channels, partitioned into 4 clients. In all three experiments, each client is equipped with 12,500 examples from the dataset, and we set a global test set of 10,000 examples. We use $E=1$, $B=500$, and employ $p=5$, $\tau=0.05$, and $c=1.01$. Each VGG model is set with a learning rate of $0.002$. We compare the performance of $5$-FLARE with FFL, EF21, FedProx Error Correction. As shown in Fig.\ref{VGG}, $5$-FLARE demonstrates faster convergence and sparsity compared to all compared methods, with superior overall performances. On VGG16, after 1000 rounds, all other methods fail to achieve a test accuracy over $0.3$, while $5$-FLARE merges with FedAvg and reaches a test accuracy of 0.5 after 410 rounds. On VGG11, FLARE achieves a test accuracy of $0.5$ after 310 rounds, while all compared methods reach this accuracy level not sooner than 950 rounds. After 1000 rounds, FLARE, FFL, and Error Correction achieve test accuracies of 0.65 and 0.52, respectively. On VGG19, there is a significant delay for all the sparsification methods, with FLARE starting to converge at round 1300, while the others methods start around 1000 but converge much slower from this point. Specifically, FLARE reaches a test accuracy of 0.5 after 1600 rounds, while all others achieve this accuracy level only around 2200 rounds. At 3000 rounds, the test accuracy achieved by FLARE and Error Correction is 0.67 and 0.61, respectively. These experiments serve as a demonstration of FLARE on large-scale DNN models, showcasing its ability to successfully accelerate the sparse training process on complex ML tasks such as VGG models.

\subsection{Experiment 3: Text generation based on "The Complete Works of William Shakespeare"}

\begin{figure}
         \centering
         \begin{subfigure}{0.4\textwidth}
            \includegraphics[width=\textwidth]{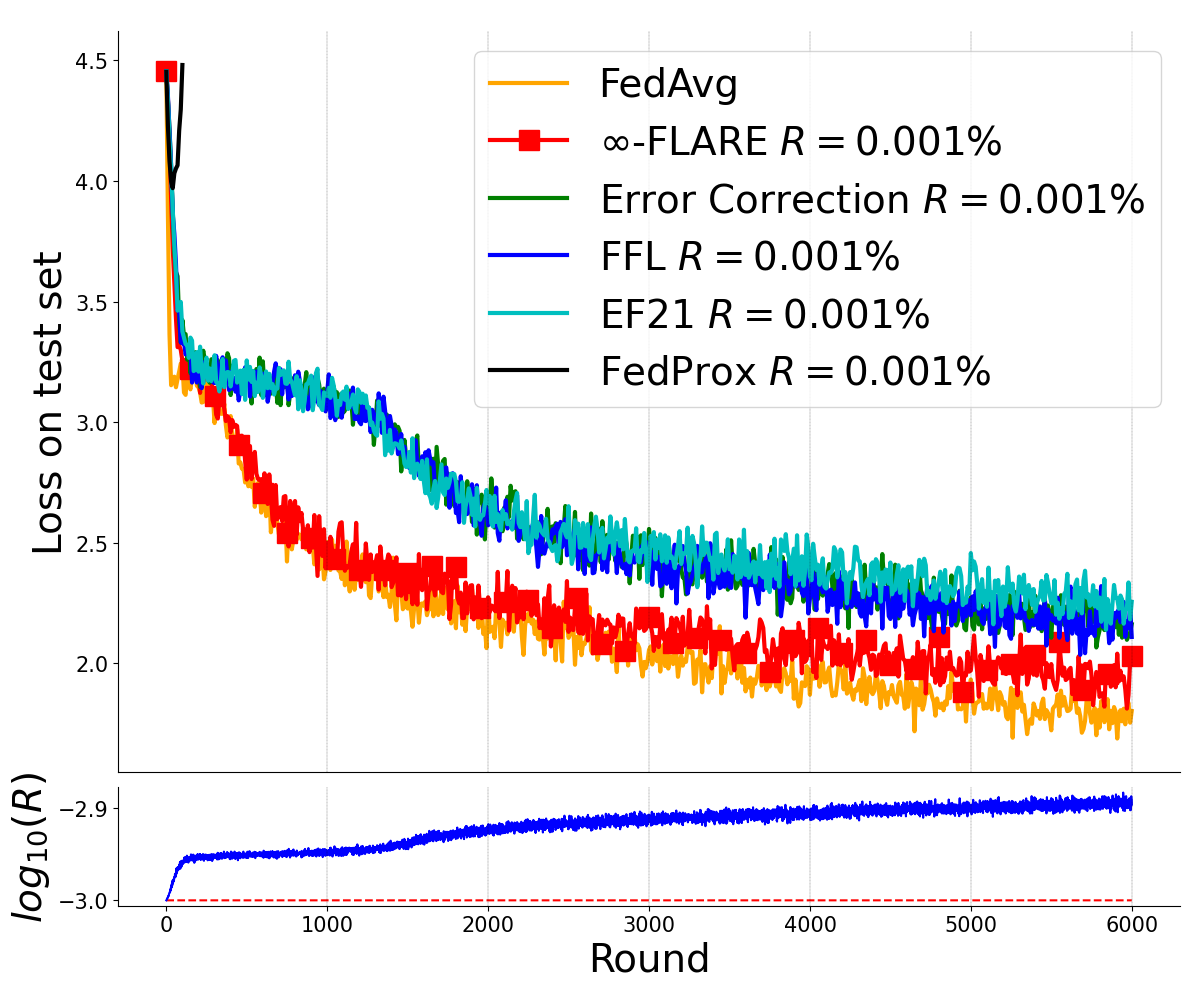}
         \end{subfigure}       
    \caption{The evaluated loss on the test set for "The Complete Works of William Shakespeare" experiment. FLARE is set with $R=0.001\%$, $p=\infty$, $\tau=0.05$, $c=1.01$. 10 clients are chosen, and we train for $E=1$ and batch size of 8. FLARE is compared with uncompressed FedAvg (benchmark for performance), Error Correction, EF21, FedProx and FFL methods.}
    \label{GRU}
    \end{figure}

We have conducted text generation experiments based on "The Complete Works of William Shakespeare" \cite{caldas2019leaf}, introduced in \cite{mcmahan2017communication} to simulate a real-world FL task. In this setting, each client represents a different character from Shakespeare's plays, thus providing unique local text data. This setup creates a realistic FL environment with unbalanced, non-IID data. The model is trained to predict the next character in the text sequence, following the setup in \cite{mcmahan2017communication}, with the following parameters. The FL system consists of $10$ clients, and utilizes a large GRU architecture with $4,022,850$ parameters. Our configuration aligns with the settings described in \cite{Tensorflow_text_generation} for $10$ clients, though our model was trained from scratch with a learning rate of $0.05$. For evaluation, we used the same test set proposed by \cite{caldas2019leaf}. In these experiments, each client operated with a batch size of $8$, $E=1$ and $R=0.001\%$. FLARE was tested with $\tau=0.05$, $c=1.01$ and $p=\infty$ (i.e. regularization is used every optimization step). As in all other experiments, we set the masking coefficient $a_0$ to the median of the accumulated absolute error values. Fig. \ref{GRU} displays the loss curves on the test set for each method evaluated. Notably, FLARE demonstrates a significant performance advantage over the other algorithms, following FedAvg without any performance degradation.

\subsection{Experiment 4: Client unavailability}
\begin{figure*}[h!]
         \centering
         \begin{subfigure}{0.25\textwidth}
            \includegraphics[width=\textwidth]{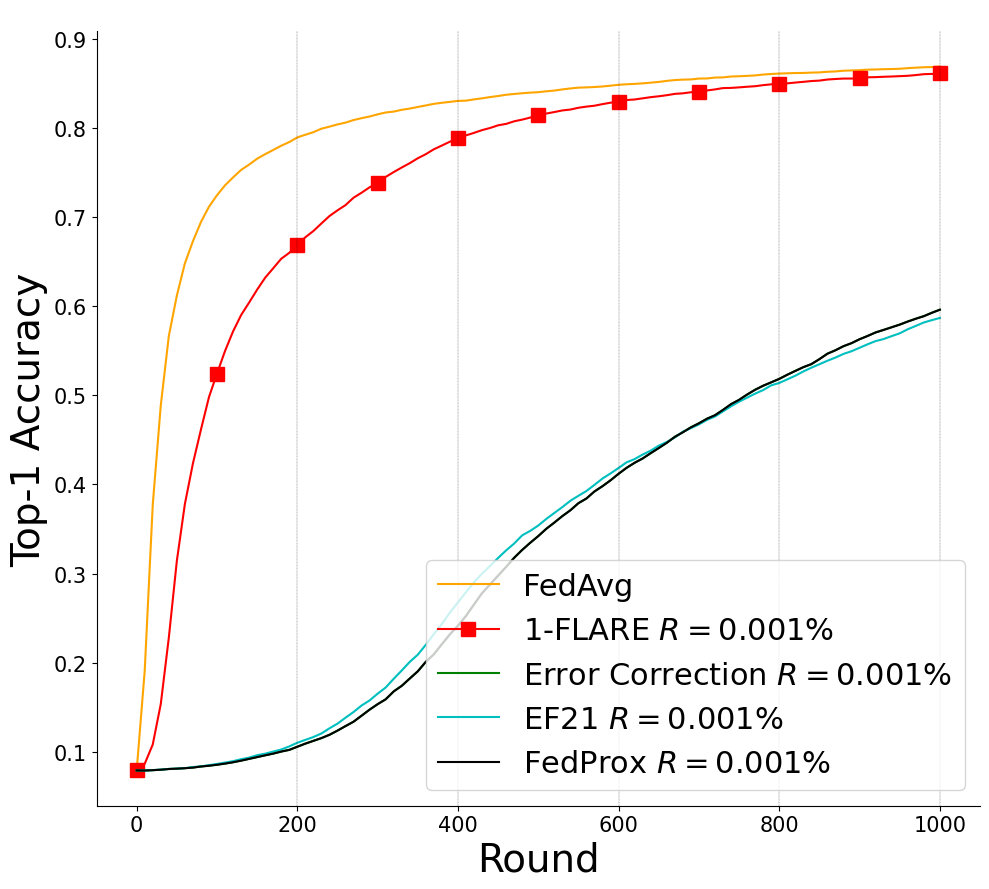}
            \caption{FC with 7 available clients}
         \end{subfigure}
         \begin{subfigure}{0.25\textwidth}
            \includegraphics[width=\textwidth]{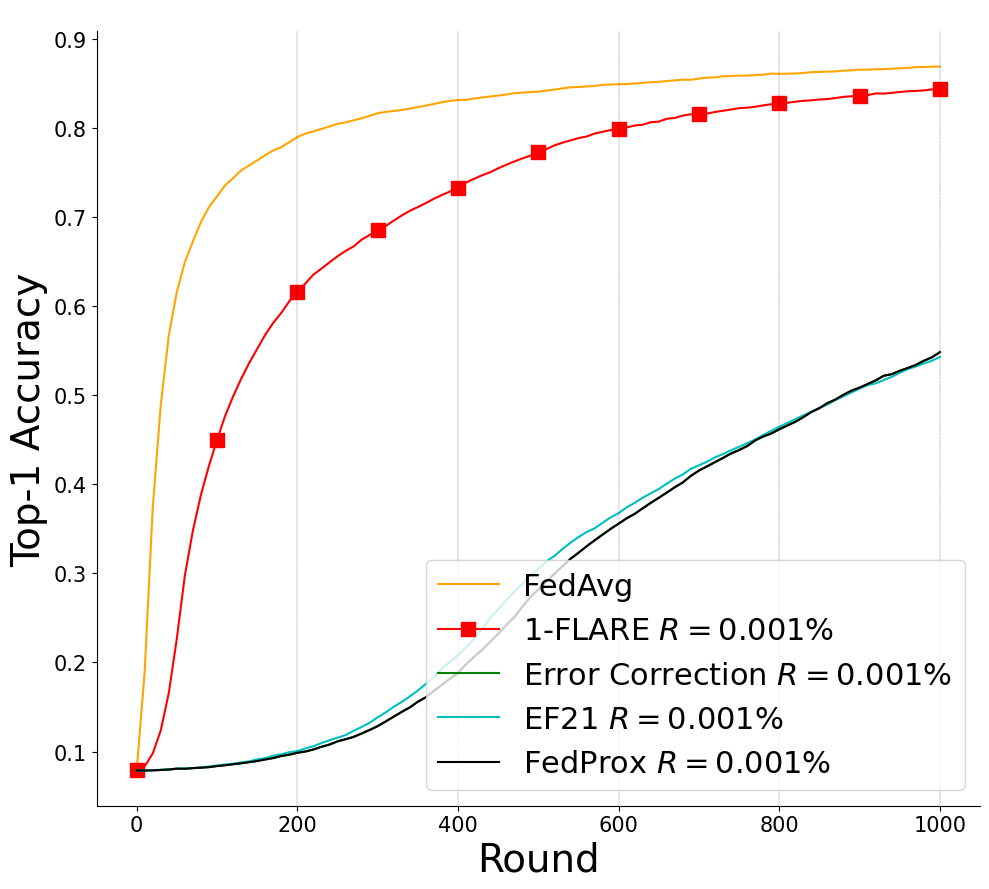}
            \caption{FC with 5 available clients}
         \end{subfigure}          
         \centering
         \begin{subfigure}{0.25\textwidth}
            \includegraphics[width=\textwidth]{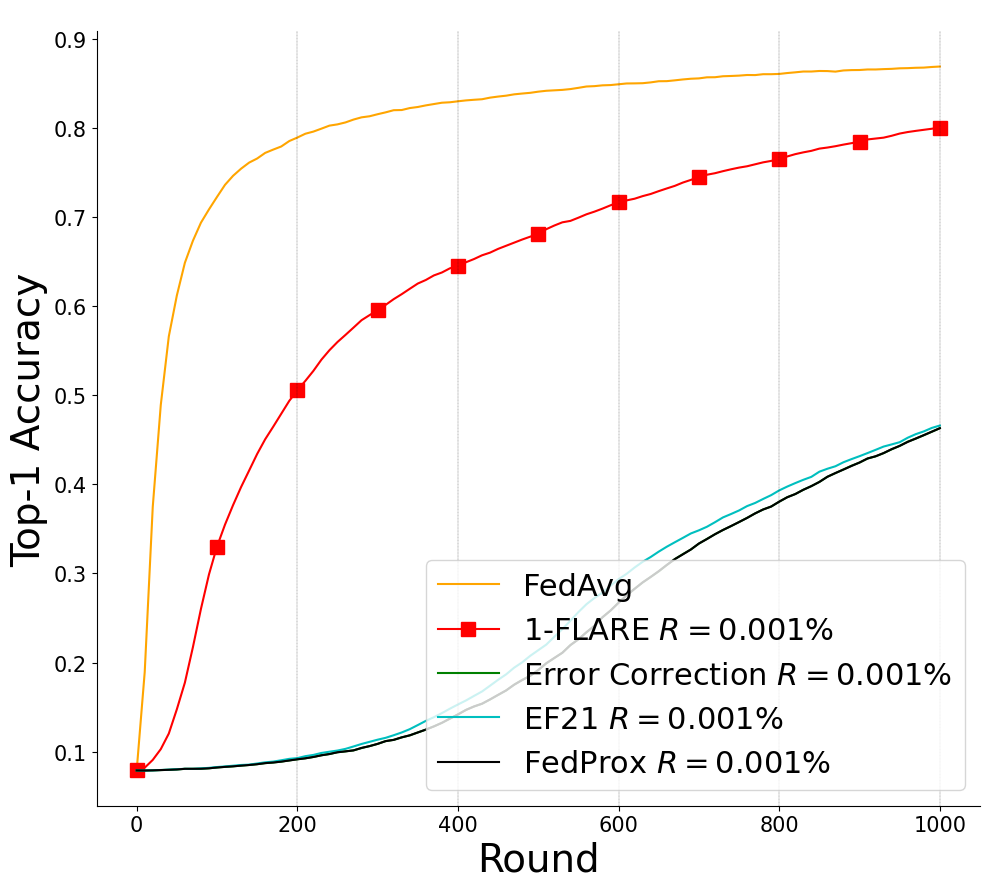}
            \caption{FC with 3 available clients}
         \end{subfigure}
         \centering
         \begin{subfigure}{0.25\textwidth}
            \includegraphics[width=\textwidth]{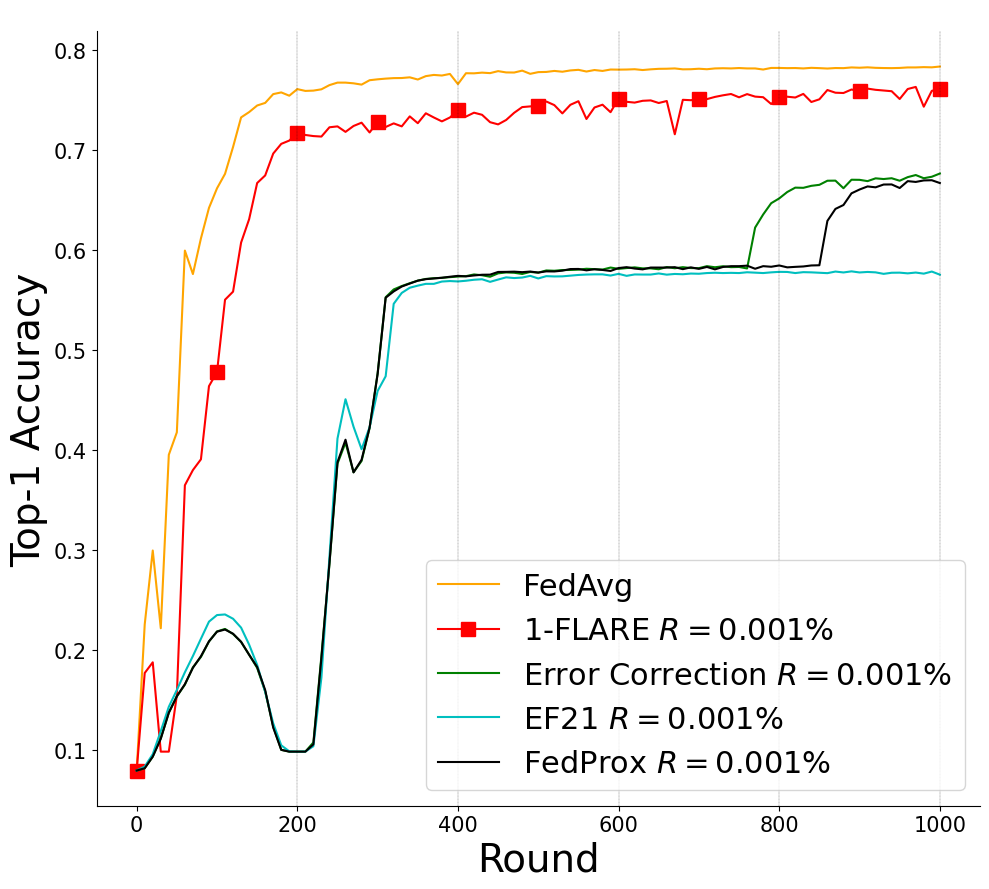}
            \caption{CNN with 7 available clients}
         \end{subfigure}
         \begin{subfigure}{0.25\textwidth}
            \includegraphics[width=\textwidth]{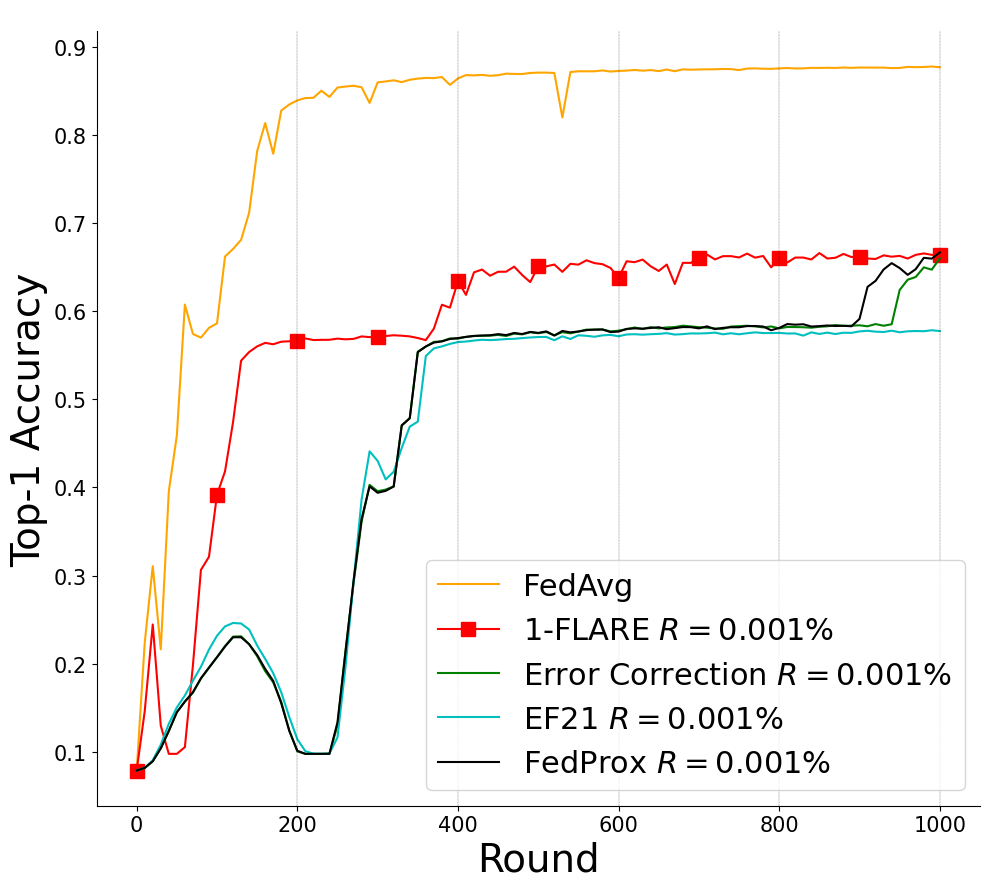}
            \caption{CNN with 5 available clients}
         \end{subfigure}          
         \centering
         \begin{subfigure}{0.25\textwidth}
            \includegraphics[width=\textwidth]{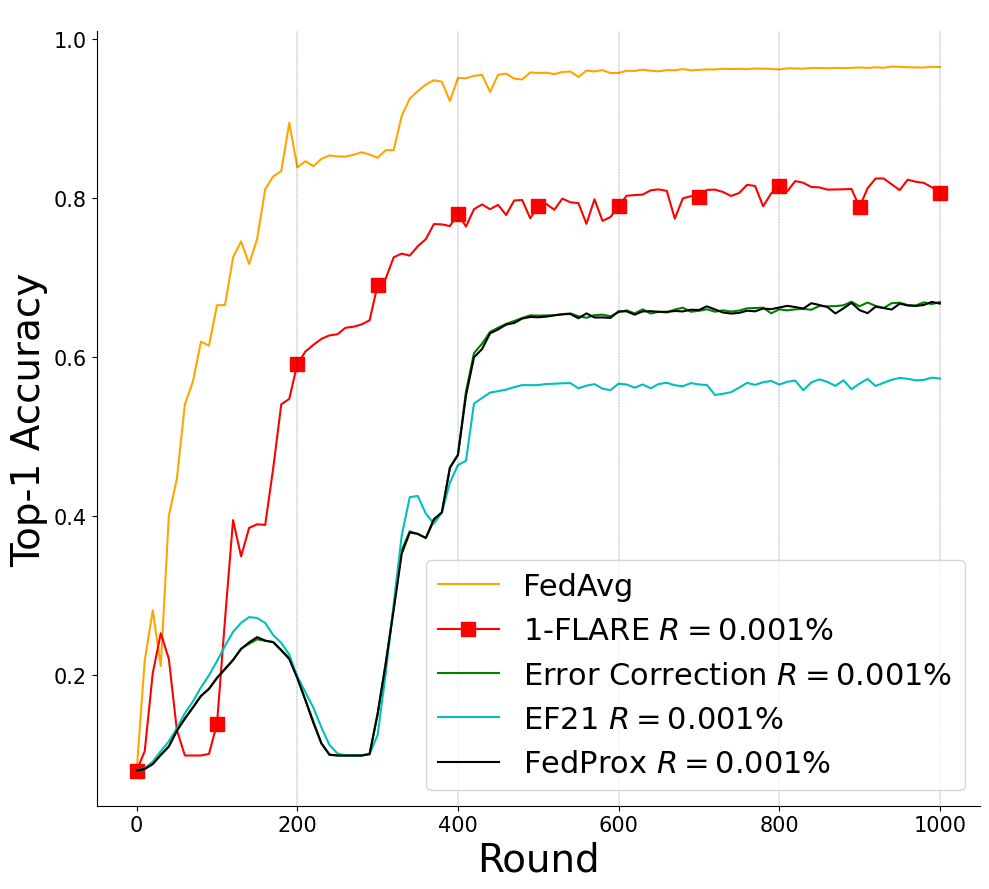}
            \caption{CNN with 3 available clients}
         \end{subfigure}
    \caption{The test accuracy performance for client unavailability compared for FC and CNN models on digit MNIST for $E=1$ (all other settings were identical to those in the FC experiment). In each round, 7, 5, or 3 out of the 10 clients were randomly selected.}
    \label{client_unavailability_FC_CNN}
    \end{figure*}

Next, we evaluate how client unavailability affects the convergence and performance of our method. Specifically, we have conducted the experiments with the FC and CNN architectures where, out of 10 total clients, only 7, 5, or 3 randomly selected clients were available during each communication round. For each scenario, we tested with $E=1$. The results, displayed in Figure \ref{client_unavailability_FC_CNN}, show that while performance degradation occurs with fewer available clients across all algorithms, as expected, the proposed FLARE algorithm continues to show superior performance compared to the other methods.

\subsection{Experiment 5: Imbalanced MNIST Dataset with the FC Model}

\begin{figure}[h]
    \centering
         \begin{subfigure}{0.24\textwidth}
            \includegraphics[width=\textwidth]{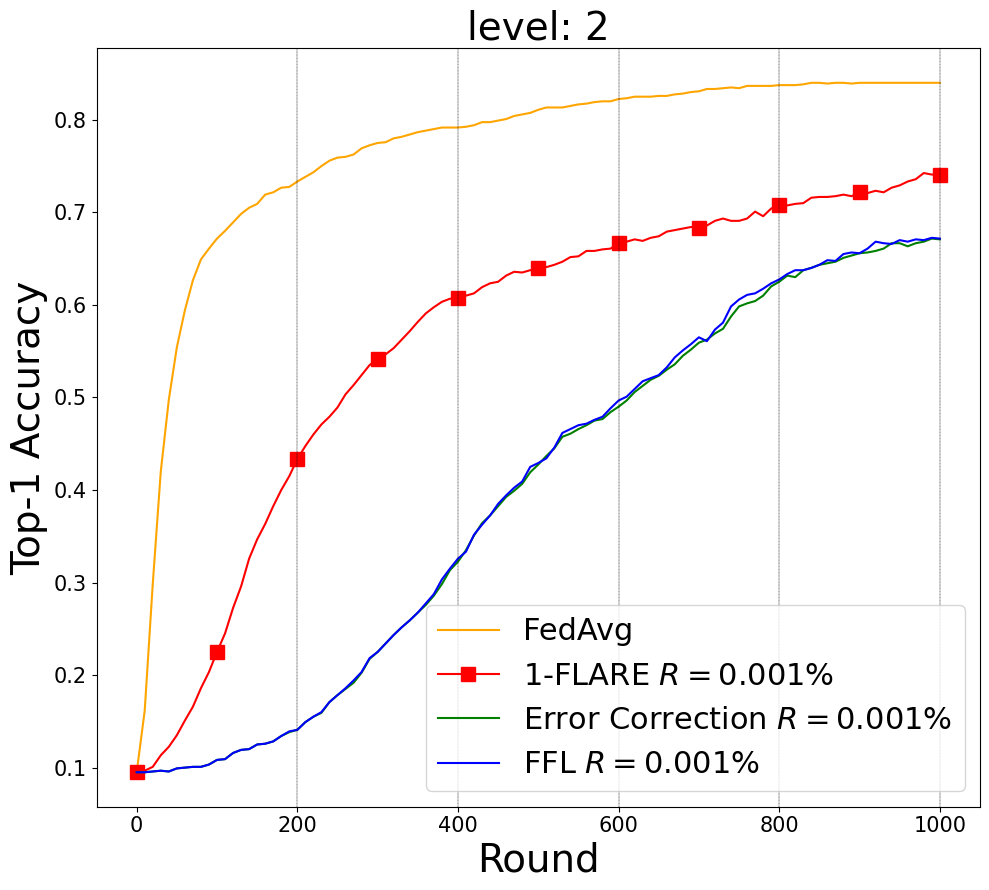}
         \end{subfigure}
         \begin{subfigure}{0.24\textwidth}
            \includegraphics[width=\textwidth]{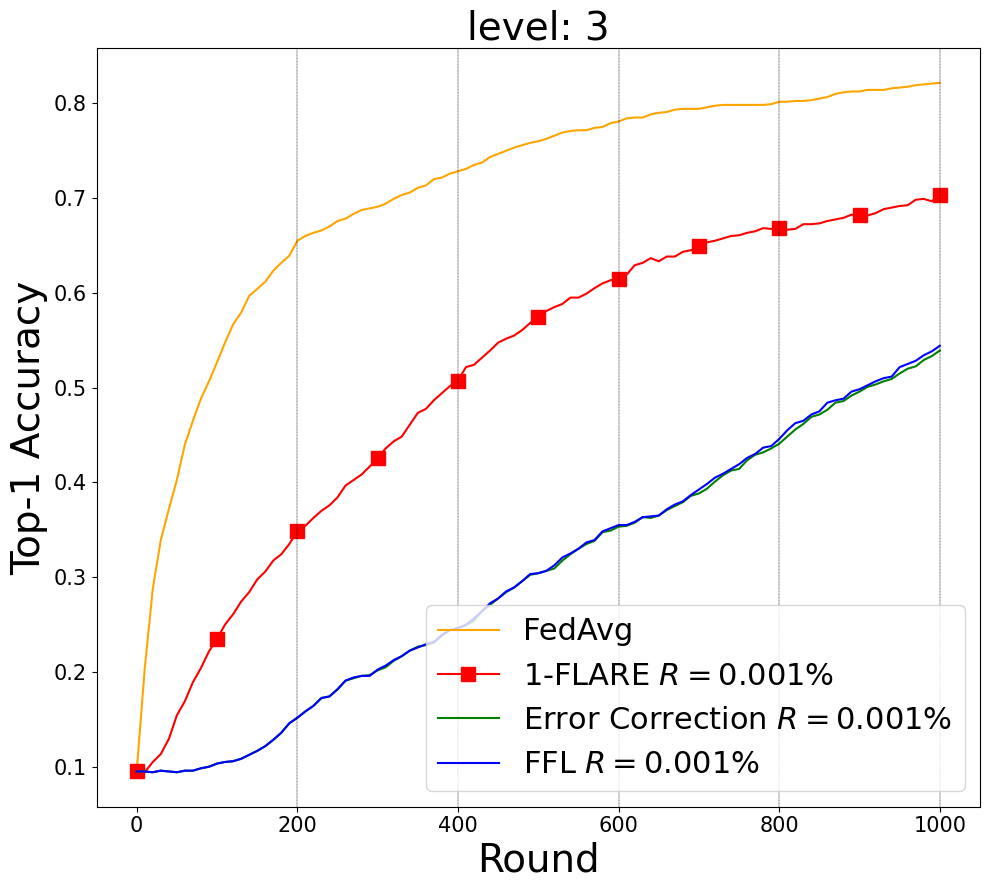}
         \end{subfigure}
         \begin{subfigure}{0.24\textwidth}
            \includegraphics[width=\textwidth]{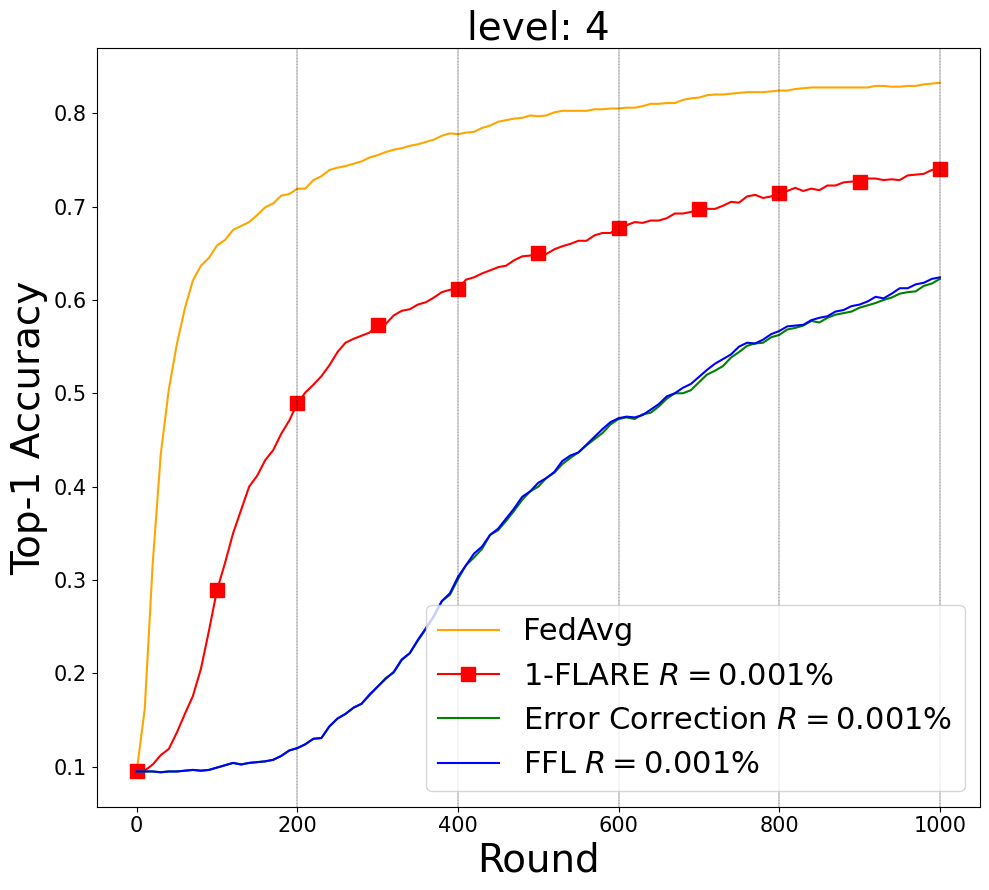}
         \end{subfigure} 
          \begin{subfigure}{0.24\textwidth}
            \includegraphics[width=\textwidth]{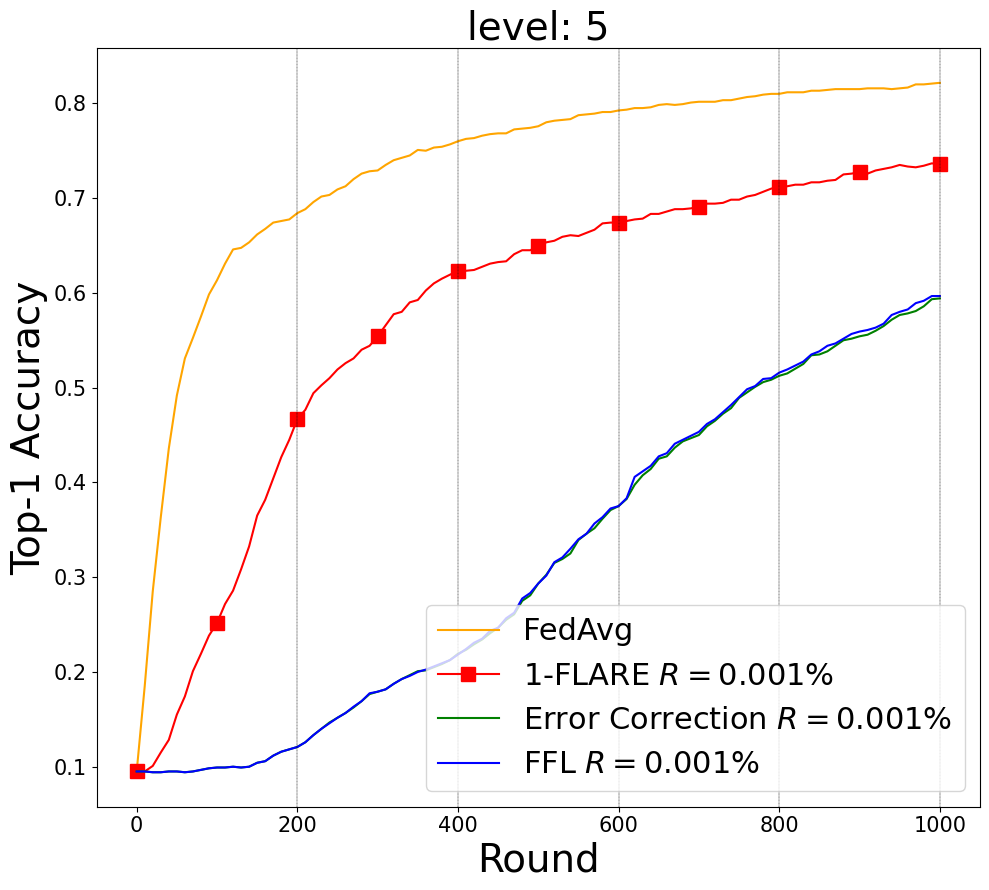}
         \end{subfigure} 
    \caption{The test accuracy performance is compared for FC model on MNIST of four different levels of imbalance in data distributions. $1$-FLARE is compared with uncompressed FedAvg (benchmark for performance), Error Correction and FFL methods. It is evident that $1$-FLARE significantly outperforms Error Correction and FFL in all cases.}
    \label{Non-iid}
    \end{figure}

In this experiment, we assess the performance of $p$-FLARE in imbalanced scenarios of class distributions using the FC model. We partition the MNIST dataset into four distinct configurations: Distributing 1200 digit examples to 5 clients, with each client exclusively receiving examples corresponding to 2, 3, 4, or 5 different digit labels, while maintaining the test set unchanged. This setup introduces varying degrees of class imbalance, with each client possessing only 2 different digits (referred to as imbalance level 2), and dataset labels among clients do not overlap. The FC model is tested for $E=1$ and $R=0.001\%$. The Top-1 test accuracy results are presented in Fig.\ref{Non-iid}. For the 2-level imbalance, after 1000 rounds, FedAvg achieves a test accuracy of 0.83, while $1$-FLARE achieves 0.74, and FFL and Error Correction achieve 0.67. For the 3-level imbalance, FedAvg achieves a test accuracy of 0.82, while $1$-FLARE achieves 0.7, and FFL and Error Correction achieve 0.54. In the case of 4-level imbalance, FedAvg achieves a test accuracy of 0.84, while $1$-FLARE achieves 0.74, and FFL and Error Correction achieve 0.62. Finally, for the 5-level imbalance, FedAvg achieves a test accuracy of 0.83, while $1$-FLARE achieves 0.73, and FFL and Error Correction achieve 0.59. 


\subsection{Experiment 6: Evaluation with reduced sparsity}
\begin{figure*}[t!]
         \centering
         \begin{subfigure}{0.25\textwidth}
            \includegraphics[width=\textwidth]{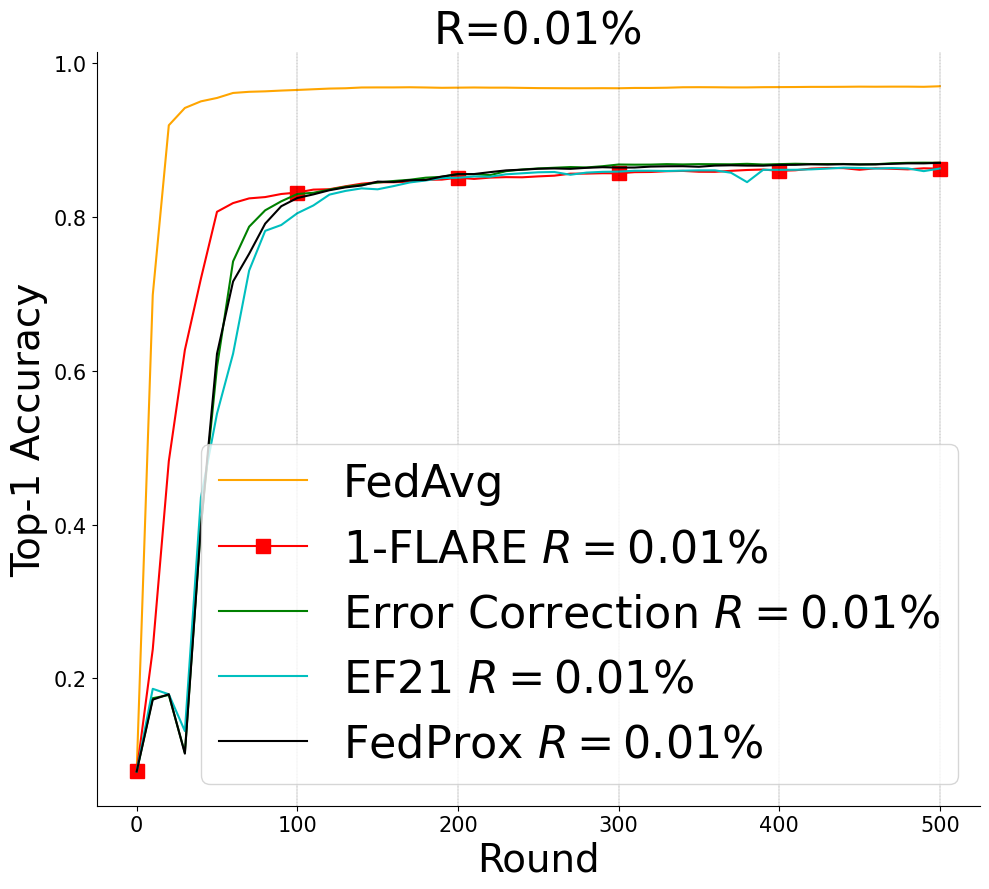}
            \caption{$E=4$}
         \end{subfigure}
         \begin{subfigure}{0.25\textwidth}
            \includegraphics[width=\textwidth]{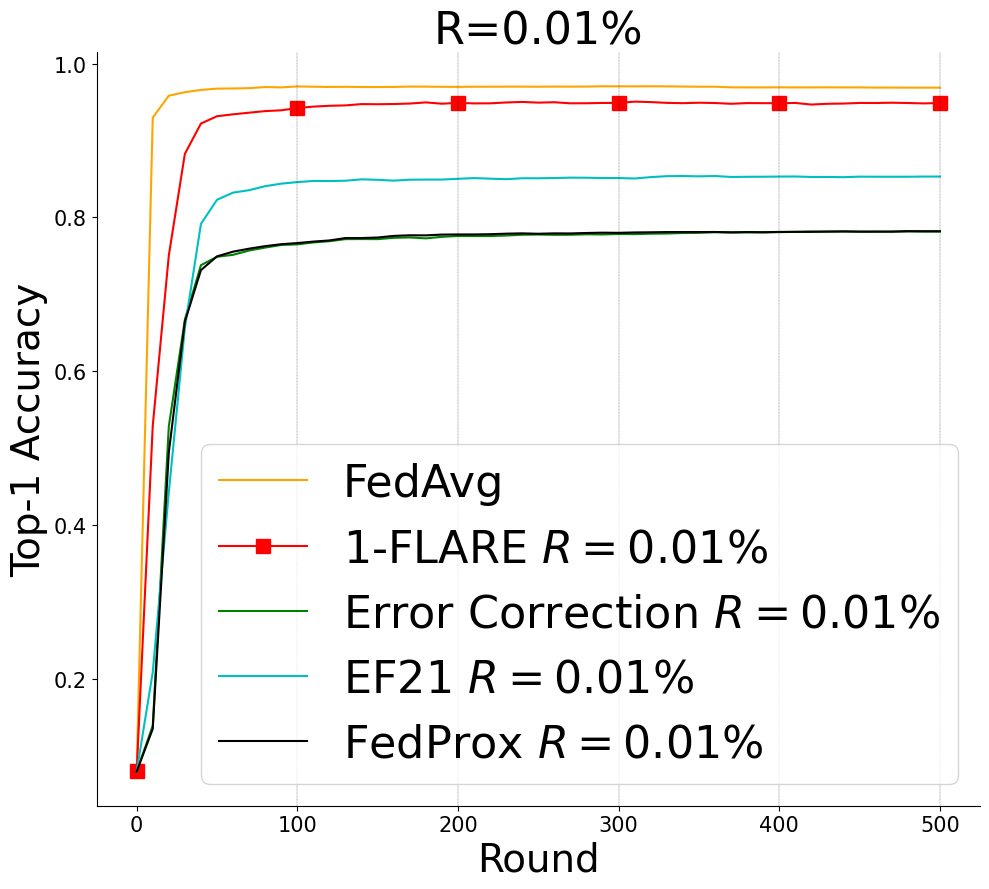}
            \caption{$E=8$}
         \end{subfigure}          
         \centering
         \begin{subfigure}{0.25\textwidth}
            \includegraphics[width=\textwidth]{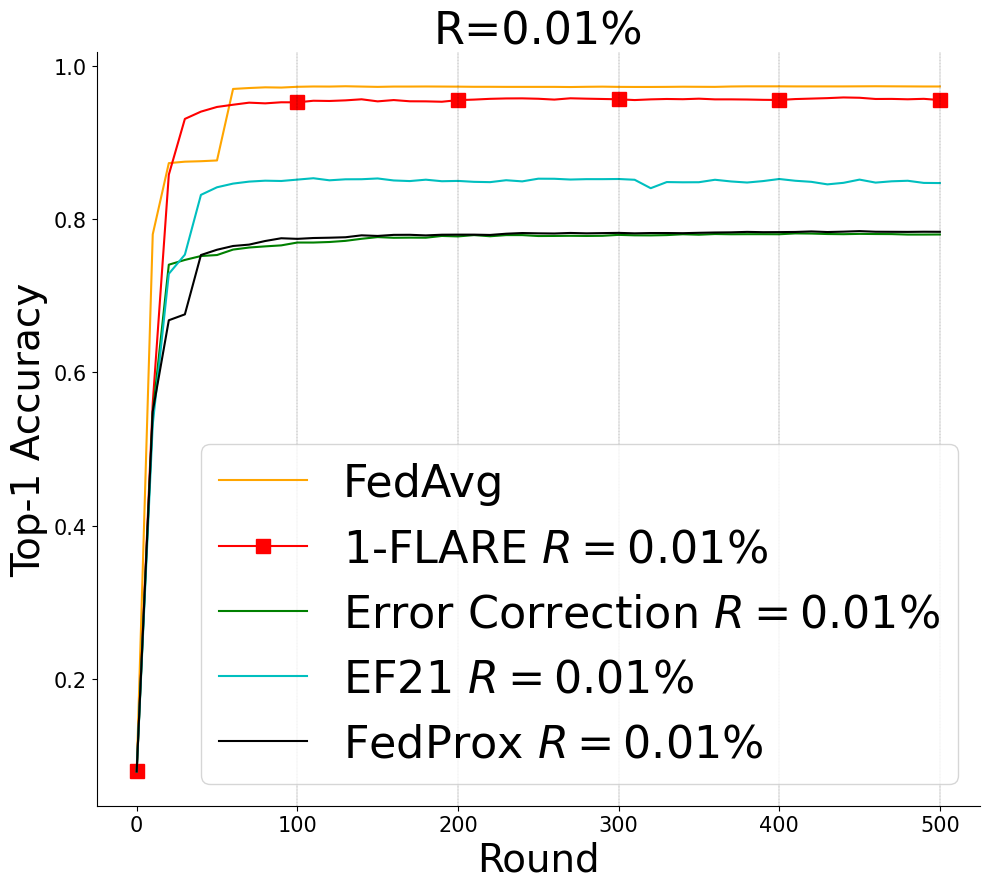}
            \caption{$E=16$}
         \end{subfigure}
         \centering
         \begin{subfigure}{0.25\textwidth}
            \includegraphics[width=\textwidth]{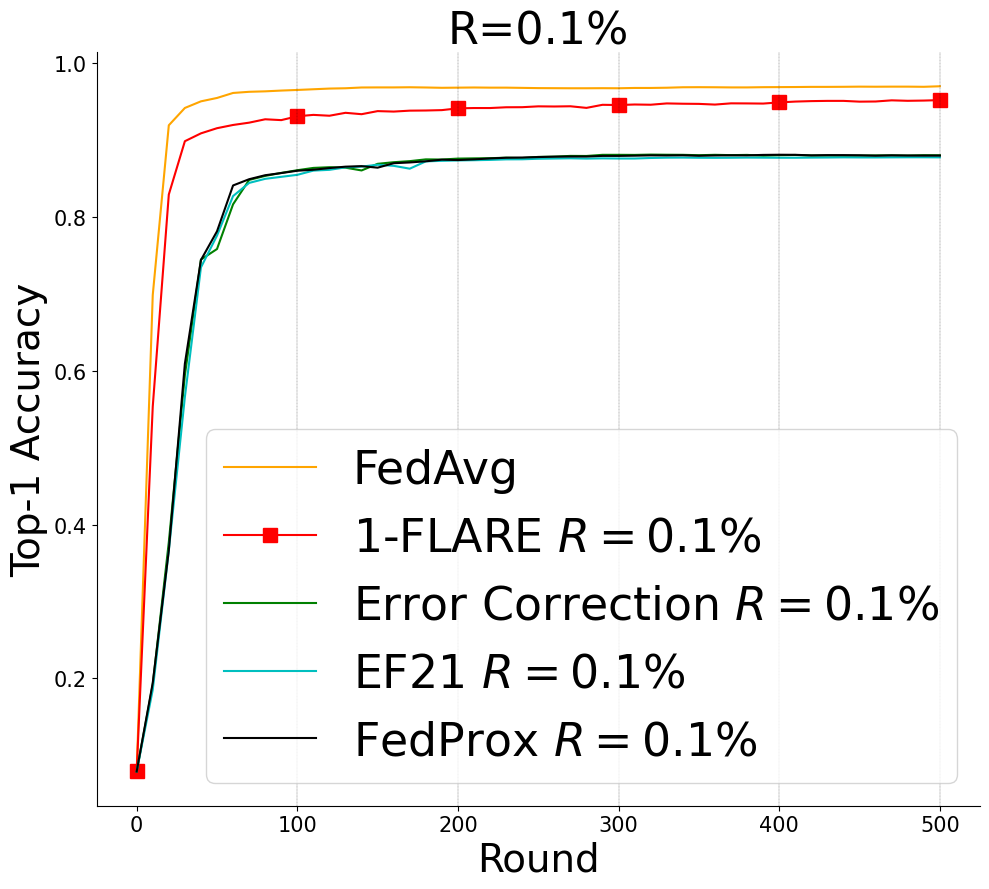}
            \caption{$E=4$}
         \end{subfigure}
         \begin{subfigure}{0.25\textwidth}
            \includegraphics[width=\textwidth]{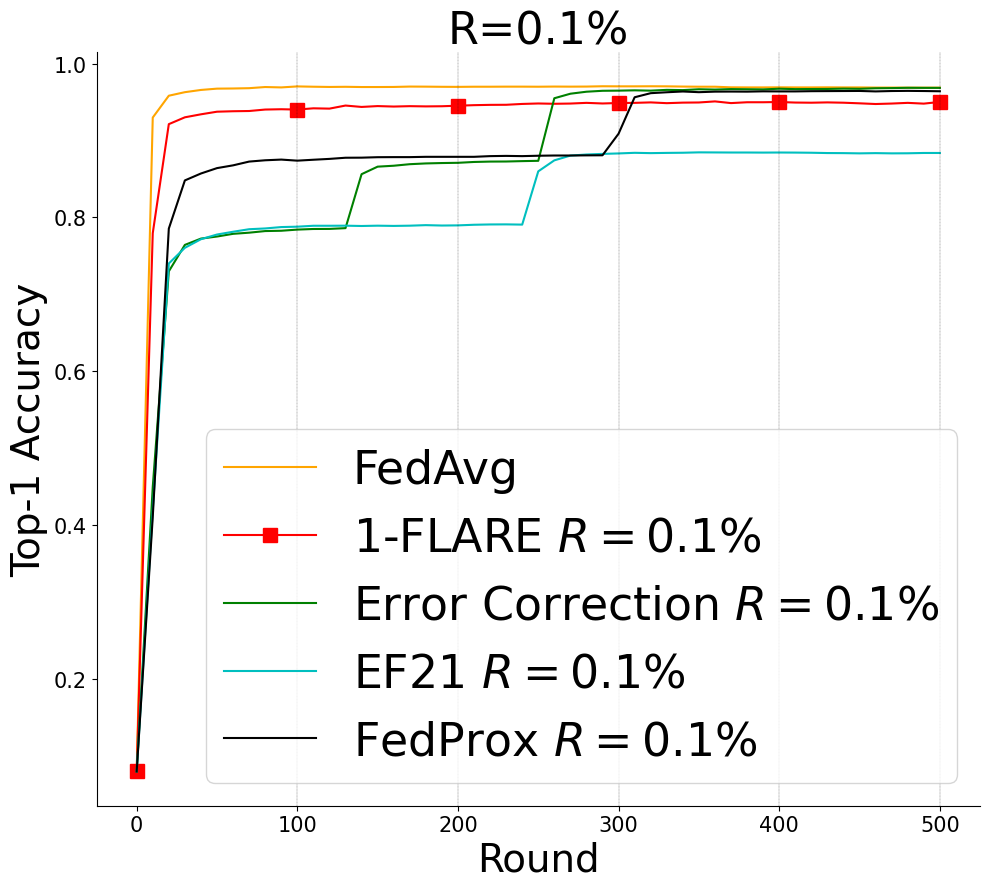}
            \caption{$E=8$}
         \end{subfigure}          
         \centering
         \begin{subfigure}{0.25\textwidth}
            \includegraphics[width=\textwidth]{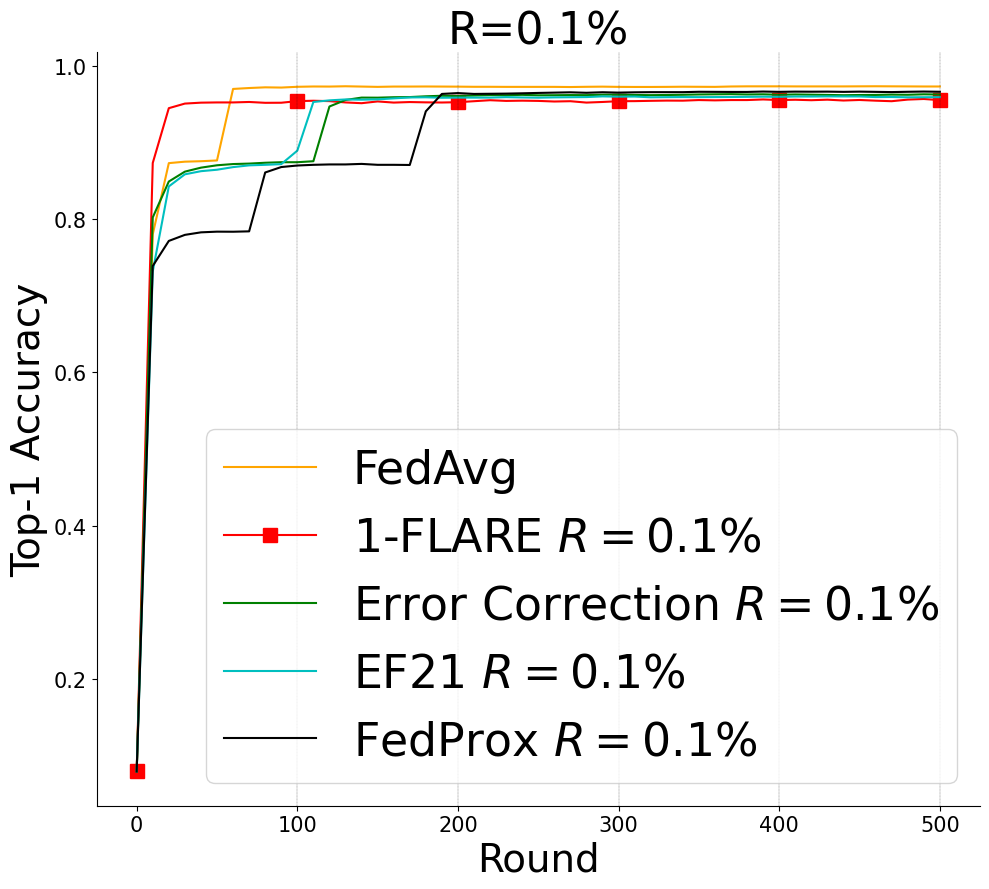}
            \caption{$E=16$}
         \end{subfigure}         
    \caption{The test accuracy performance is compared for CNN model on digit MNIST with $R=0.01\%$ and $R=0.1\%$, $p=1$, $\tau=0.05$, $c=1.1$, 10 clients with $B=\infty$ and $E=4,8,16$. $p$-FLARE is compared with uncompressed FedAvg (benchmark for performance), Error Correction, EF21 and FedProx methods.}
    \label{0_01&0_1}
    \end{figure*}

Additional experiments are conducted to evaluate the performance of our proposed method at lower sparsity levels. Specifically, we included scenarios with $R=0.01\%$ and $R=0.1\%$ sparsity for the CNN experiments, as shown in Figures \ref{0_01&0_1}. Our findings indicate that as sparsity decreases, FLARE maintains superior performance compared to other error feedback algorithms. However, as sparsity increases, FLARE's performance significantly surpasses the others, which is especially advantageous for dimensionality reduction applications. This trend is further validated by the theoretical performance bounds developed in Section IV, demonstrating that FLARE achieves a superior scaling rate with respect to sparsity.

\section{Conclusion}

Error Correction has proven to be a promising technique for alleviating communication overhead in FL systems. However, when sparsity is taken to the extreme, it struggles with staled updates, posing a challenge to maintaining model performance. In Error Correction, clients only exchange the most significant changes, retaining the rest locally for accumulation and releasing them for transmission once they accumulate sufficiently.
This paper presented the FLARE algorithm, a significant improvement over methods based on Error Correction. FLARE allows pushing FL communication sparsity to the extreme, reducing its overhead by addressing the staleness effect and mitigating its influence. FLARE only requires a modification to the loss function, making use of accumulated values. Our theoretical analysis demonstrates that FLARE's regularized error feedback not only matches state-of-the-art techniques in terms of convergence rate with time, but also achieves significant improvements in scalability with sparsity parameter. Through extensive experiments, we systematically evaluate the performance of FLARE. Specifically, FLARE achieved a sparsity level of $R=0.001\%$, where vanilla methods failed to converge, and in some cases, even when sparsity is set to be one order larger, FLARE demonstrated superior performance. These findings establish the FLARE algorithm as a highly promising solution for practical implementation in FL systems constrained by communication limitations.

\appendix

\section{Proofs}

\subsection{Proof of Theorem~\ref{Theorem1}}
\noindent
To streamline the notations throughout the proof, we omit the bar above vectors where the context is clear. To prove the theorem, a bound for the accumulated error is derived first. In the case where clients perform one optimization step, the model deltas $\Delta{w}^i_{t}$ produced by each client are the negative gradients. Then, the updates translate to  ${A}^i_{t} \leftarrow \gamma \nabla F^i_{t} + {A}^i_{t-1}$, and aggregating $-{S}^i_t + {w}_{t-1}$. 
Recall that with FLARE the objective becomes:
\begin{align}
    \tilde{F}({w};{w}_t,{A}_t) = F({w}) + \frac{\tau}{2} || {w} - ({w}_t + {A}_t)||^2.
\end{align}
By defining $\tilde{g}_{t}$ as $\nabla  \tilde{F}({w}_t;{w}_t,{A}_t)$ and $ g_{t} $ as $\nabla F(w_t)$ we have $\tilde{g}_{t} = g_{t} - \tau A_t$.
\noindent
The update equation for $A_{t+1}$ can now be written as: 
\begin{align}
A_{t+1} = \gamma \tilde{g}_{t} + A_{t}-\mathcal{C}(\tilde{g}_{t} + A_{t}).
\label{A_t relation}
\end{align}
Using Assumption~\ref{assumption1}, we have:
$$
\begin{aligned}
\left\|A_{t+1}\right\|^2 & =\left\|\gamma \tilde{g}_{t} + A_{t}-\mathcal{C}(\gamma \tilde{g}_{t} + A_{t})\right\|^2 \leq \\ 
                         &(1-\delta)\left\|\gamma \tilde{g}_{t} + A_{t}\right\|^2.
\end{aligned}
$$
Using the definition of $\tilde{g}_{t}$, we have:

\noindent
$$
\begin{aligned}
\quad\left\|A_{t+1}\right\|^2 \leq &  (1-\delta)\left\|\gamma \tilde{g}_{t}+A_t\right\|^2  = \\
                                   & (1-\delta) \cdot \| \gamma g_{t}+\left(1-\gamma \tau) A_t \|^2.\right .
\end{aligned}
$$
Next, we apply the inequality $\|\mathbf{a}+\mathbf{b}\|^2 \leq(1+\gamma)\|\mathbf{a}\|^2+\left(1+\gamma^{-1}\right)\|\mathbf{b}\|^2$, and for any $\eta > 0$ we have:
$$
\begin{aligned}
& \left\| \gamma g_{t} + (1-\gamma \tau) A_t \right\|^2 \leqslant \\
& \gamma^2 \cdot(1+\frac{1}{\eta}) \cdot\left\| g_{t}\right\|^2+(1+\eta) \cdot(1-\gamma \tau)^2 \cdot\left\| A_t \right\|^2.
\end{aligned}
$$
We obtain a recursive relation:
$$
\begin{aligned}
\left\|A_{t+1}\right\|^2 \leqslant & (1-\delta) \cdot (1+\eta) \cdot(1-\gamma \tau)^2 \cdot\left\|A_t\right\|^2 + \\
                                   &(1-\delta) \cdot \gamma^2 \cdot(1+\frac{1}{\eta}) \cdot\left\|g_{t}\right\|^2.
\end{aligned}
$$

Solving the recursion (where $A_0 = 0$) yields:
$$
\begin{aligned}
& \left\{\begin{array}{l}
A_1<a A_0+b g_0=b g_0 \\
A_2<a A_1+b g_1<a b g_0+b g_1 \\
A_3<a A_2+b g_2<a^2 b g_0+a b g_1+b g_2
\end{array}\right\}.
\end{aligned}
$$ 
Then, we get 
$$
\begin{aligned}
& \left\|A_t\right\|^2  \leq 
(1-\delta) \gamma^2 \left(1+\frac{1}{\eta}\right) \\
& \hspace{0.5cm}\times \sum_{i=0}^{t-1}\left[(1+\eta)(1-\gamma \tau)^2(1-\delta)\right]^{t-i-1} \cdot\left\|g_{i}\right\|^2 .
\end{aligned}
$$
Using assumption~\ref{assumption3} and choosing $\eta = \dfrac{\delta}{2(1-\delta)}$ yield $(1-\delta)(1+\eta)=1-\dfrac{\delta}{2}$ and $1+\dfrac{1}{\eta} < \dfrac{2}{\delta}$. Then, we get:
$$
\begin{aligned}
     & \mathbb{E} \left\|A_t\right\|^2  \leq  \gamma^2  \left(1+\frac{1}{\eta}\right)  (1-\delta) \\
     & \hspace{5ex} \times \sum_{i=0}^{t-1}\left[(1+\eta)(1-\gamma \tau)^2(1-\delta)\right]^{t-i-1} \mathbb{E} \left\|g_{i}\right\|^2 \\
     &  \leq \gamma^2 \left(1+\frac{1}{\eta}\right) (1-\delta) \sigma^2 \sum_{i=0}^{\infty}\left[(1+\eta)(1-\gamma \tau)^2(1-\delta)\right]^{i}. \\   
\end{aligned}
$$
Then,
$$
\begin{aligned}
     & \mathbb{E} \left\|A_t\right\|^2  \leq 
     \dfrac{\gamma^2  \left(1+\frac{1}{\eta}\right)  (1-\delta) \cdot \sigma^2}{1- \left(1- \dfrac{\delta}{2} \right) (1-\gamma \tau)^2}  \\
     & \hspace{10ex} \leq\dfrac{2 \gamma^2 (1-\delta)\sigma^2}{\delta} \cdot \dfrac{1}{1 - (1-\dfrac{\delta}{2}) (1-\gamma \tau)^2} \\
     & = \dfrac{2 \gamma^2 (1-\delta)\sigma^2}{\delta} \cdot \dfrac{\nicefrac{2}{\delta}}{\dfrac{2}{\delta} \left( \dfrac{\delta}{2} + \gamma \tau (2-\delta) + \gamma^2\tau^2 (\dfrac{\delta}{2}-1) \right)} \\
     & = \dfrac{4 \gamma^2 (1-\delta)\sigma^2}{\delta^2} \cdot \dfrac{1}{1+\dfrac{2}{\delta}\left( \gamma \tau(2-\delta) +\gamma^2 \tau^2 (\dfrac{\delta}{2} - 1)\right)}\\
     & = \dfrac{4 \gamma^2 (1-\delta)\sigma^2}{\delta^2} \cdot \dfrac{1}{1+p(\tau)}.
\end{aligned}
$$

From this point forward, we define $\hat{w_t}$ as $\hat{w}_t =w_t-A_t$ and $\hat{w}_{t+1} =w_{t+1}-A_{t+1}$. Since $w_{t+1} = w_{t}-\mathcal{C}(\gamma \tilde{g}_{t} + A_{t})$ (due to aggregation step), using \eqref{A_t relation} yields:
$$
\begin{aligned}
\hat{w}_{t+1}  & = w_{t+1}-A_{t+1} = \\
& w_{t}-\mathcal{C}(\gamma \tilde{g}_{t} + A_{t})-\left(\gamma \tilde{g}_{t} + A_{t}-\mathcal{C}(\tilde{g}_{t} + A_{t})\right) \\
               & = w_t-\gamma \tilde{g}_{t} -  A_{t} = \hat{w}_t-\gamma \tilde{g}_t.
\end{aligned}
$$
Using the accumulated error bound for $\mathbb{E} \left\|A_t\right\|$ and C.S inequality, we have:
$$
\begin{aligned}
& \mathbb{E}\left[\left\|\hat{w}_{t+1}-w^*\right\|^2\right]=\mathbb{E}\left[\left\|\hat{w}_t-\gamma \tilde{g}_t-w^*\right\|^2\right] \\
& =\left\|\hat{w}_t-w^*\right\|^2+\gamma^2 \mathbb{E}\left[\left\|\tilde{g}_t\right\|^2\right]-2 \gamma\left\langle  \mathbb{E}_t\left[\tilde{g}_t\right], \hat{w}_t-w^*\right\rangle \\
& \leqslant\left\|\hat{w}_t-w^*\right\|^2+\gamma^2 \sigma^2-2 \gamma\left\langle\mathbb{E}_t\left[\tilde{g}_t\right], w_t-\hat{w}_t^*\right\rangle + \\
\end{aligned}
$$
$$
\begin{aligned}
& \hspace{5ex} + 2 \gamma\left\langle\mathbb{E}_t\left[\tilde{g}_t\right], w_t-\hat{w}_t\right\rangle \\
& =\left\|\hat{w}_t-w^*\right\|^2+\gamma^2 \sigma^2-2 \gamma\left\langle\mathbb{E}_t\left[\tilde{g}_t\right], w_t-w^*\right\rangle + \\
& \hspace{5ex} + 2 \gamma\left\langle\mathbb{E}_t\left[\tilde{g}_t\right], A_t\right\rangle  \\
& \leqslant\left\|\hat{w}_t-w^*\right\|^2+\gamma^2 \sigma^2 - 2 \gamma\left\langle\mathbb{E}_t\left[\tilde{g}_t\right], w_t-w_w^*\right\rangle +\\
& \hspace{5ex} + 2 \gamma \mathbb{E}_t\left[\left\|\tilde{g}_t\right\|\right]  \mathbb{E}_t\left[\left\|A_t\right\|\right] \\
& \leqslant\left\|\hat{w}_t-w^*\right\|^2+\gamma^2 \sigma^2 - 2 \gamma\left\langle\mathbb{E}_t\left[\tilde{g}_t\right], w_t-w^2\right\rangle + \\
& \hspace{5ex} + 2 \gamma \sigma  \mathbb{E}_t \left[\left\|A_t\right\|\right] \\
& \leqslant\left\|\hat{w}_t-w^*\right\|^2+\gamma^2 \sigma^2 - 2 \gamma\left\langle\mathbb{E}_t\left[\tilde{g}_t\right], w_t-w^*\right\rangle + \\
& \hspace{5ex} + 4 \gamma^2 \sigma^2 \cdot \frac{\sqrt{1-\delta}}{\delta \sqrt{1+p(\tau)}}.
\end{aligned}
$$
Rearranging and summing over $T$ yields:
$$
\begin{aligned} 
& \frac{2 \gamma}{T+1} \sum_{t=0}^T \mathbb{E}\left[\left\langle\tilde{g}_t, w_t-w^*\right\rangle\right] \\
& \leq\frac{1}{T+1} \sum_{t=0}^T  \mathbb{E}\left[\left\|\hat{w}_t-w^*\right\|^2\right]-\mathbb{E}\left[\left\|\hat{w}_{t+1}-w^*\right\|^2\right] \\
& +\gamma^2 \sigma^2+4 \gamma^2 \sigma^2 \frac{\sqrt{1-\delta}}{\delta \sqrt{1+p(\tau)}}.
\end{aligned}
$$
Since  $A_0=0,\hat{w}_0=w_0$ we have:
$$
\begin{aligned} 
&\frac{1}{T+1} \sum_{t=0}^T \mathbb{E}\left[\left\langle\tilde{g}_t, w_t-w^*\right\rangle\right] \leqslant \frac{\left\|w_0 - w^*\right\|^2}{2 \gamma(T+1)} \\
\end{aligned}
$$
$$
\begin{aligned}    
&\hspace{2cm}+\gamma \sigma^2\left(\frac{1}{2}+\frac{2 \cdot \sqrt{1-\delta}}{\delta \sqrt{1+p(\tau)}}\right).
\end{aligned}
$$
Since $F$ is convex, $\tilde{F}$ is convex and therefore:
$$
\begin{aligned} 
& \frac{1}{T+1} \sum_{t=0}^T \mathbb{E}\left[\left\langle \tilde{g}_t, w_t-w^*\right\rangle\right] \geq \frac{1}{T+1} \sum_{t=0}^T \tilde{F}(w_t)-\tilde{F}(w^*) \\
& \geq \tilde{F}\left(\frac{1}{T+1} \sum_{t=0}^T w_t\right)-\tilde{F}(w^*) .
\end{aligned} 
$$
which completes the proof.\hfill $\square$
\subsection{Proof of Theorem~\ref{Theorem2}}
\noindent
We utilize the relations $\hat{w}_t =w_t-A_t$ and $\hat{w}_{t+1} =\hat{w}_t-\gamma \tilde{g}_t$ from the proof of Theorem~\ref{Theorem1} in this proof. since $F$ is L-smooth, it follows that:
$$
\begin{aligned}
    & \mathbb{E}_t\left[F\left(\hat{w}_{t +1}\right)\right] \leqslant \\
    & F\left(\hat{w}_t\right)+\left\langle\nabla F\left(\hat{w}_t\right), \mathbb{E}_t\left[\hat{w}_{t+1}-\hat{w}_t\right]\right\rangle \\
    & \hspace{10ex} +\frac{L}{2} \mathbb{E}_t\left[\left\|\hat{w}_{t+1}-\hat{w}_{t}\right\|^2\right] \\
    & =F\left(\hat{w}_t\right) - \gamma\left\langle\nabla F\left(\hat{w}_t\right), \mathbb{E}_t\left[\tilde{g}_t\right]\right\rangle + \frac{L \gamma^2}{2} \mathbb{E}_t\left[\left\|\tilde{g}_t\right\|^2\right]\\
    & =F\left(\hat{w}_t\right) - \gamma\left\langle\nabla F\left(\hat{w}_t\right), \mathbb{E}_t\left[\tilde{g}_t\right]\right\rangle + \\    
    & \hspace{10ex} + \frac{L \gamma^2}{2} \mathbb{E}_t\left[\left\|  \nabla F(w_t) - \tau A_t  \right\|^2\right]\\       
    & \leqslant F\left(\hat{w}_t\right)-\gamma \left\langle\nabla F\left(\hat{w}_t\right), \tilde{g}_t\right\rangle +\\  
    & \hspace{10ex} + \frac{L \gamma^2}{2} \sigma^2+\frac{L \gamma^2 \tau^2}{2}  \mathbb{E}_t \left[\left\|A_t\right\|^2\right] \\
\end{aligned}
$$
$$
\begin{aligned}       
    & =F\left(\hat{w}_t\right)-\gamma \left\langle\nabla F\left(\hat{w}_t\right) + \tilde{g}_t - \tilde{g}_t, \tilde{g}_t\right\rangle +\\
    & \hspace{10ex} + \frac{L \gamma^2}{2} \sigma^2+\frac{L \gamma^2 \tau^2}{2}  \mathbb{E}_t \left[\left\|A_t\right\|^2\right] \\
    & \leqslant F\left(\hat{w}_t\right)-\gamma \left\| \tilde{g}_t \right\|^2 + \gamma | \left\langle\nabla F\left(\hat{w}_t\right) - \tilde{g}_t, \tilde{g}_t\right\rangle| + \\
    &\hspace{15ex} + \frac{L \gamma^2}{2} \sigma^2+\frac{L \gamma^2 \tau^2}{2}  \mathbb{E}_t \left[\left\|A_t\right\|^2\right]. \\
    &\text{Using the mean value inequality for some $\rho > 0 $ yields:} \\
    & \mathbb{E}_t\left[F\left(\hat{w}_{t +1}\right)\right] \leqslant F\left(\hat{w}_t\right)-\gamma \left\| \tilde{g}_t \right\|^2 + \dfrac{\gamma \rho}{2} \left\| \tilde{g}_t \right\|^2 +   \\
    & \hspace{2ex} \dfrac{\gamma}{2\rho} \left\| \nabla F(\hat{w}_t) - \tilde{g}_t \right\|^2 + \frac{L \gamma^2}{2} \sigma^2+\frac{L \gamma^2 \tau^2}{2}  \mathbb{E}_t \left[\left\|A_t\right\|^2\right]. \\ 
\end{aligned}
$$
Then,
$$
\begin{aligned}    
    & \mathbb{E}_t\left[F\left(\hat{w}_{t +1}\right)\right] \leqslant 
    F\left(\hat{w}_t\right)-\gamma \left\| \tilde{g}_t \right\|^2 + \dfrac{\gamma \rho}{2} \left\| \tilde{g}_t \right\|^2   \\
    & \hspace{2ex} +\dfrac{\gamma}{2\rho} \left\| \nabla F(\hat{w}_t) - \nabla F(w_t) + \tau A_t \right\|^2 +\frac{L \gamma^2}{2} \sigma^2\\
    & \hspace{15ex} + \frac{L \gamma^2 \tau^2}{2}  \mathbb{E}_t \left[\left\|A_t\right\|^2\right] \\
    & \leqslant F\left(\hat{w}_t\right)-\gamma \left\| \tilde{g}_t \right\|^2 + \dfrac{\gamma \rho}{2} \left\| \tilde{g}_t \right\|^2  \\
    & \hspace{10ex} +\dfrac{\gamma}{2\rho} \left( \left\| \nabla F(\hat{w}_t) - \nabla F(w_t) \right\|^2 + \tau^2 \left\| A_t \right\|^2 \right)   \\
    & \hspace{15ex} +\frac{L \gamma^2}{2} \sigma^2+\frac{L \gamma^2 \tau^2}{2}  \mathbb{E}_t \left[\left\|A_t\right\|^2\right] \\
    & = F\left(\hat{w}_t\right)-\gamma \left(1-\dfrac{\rho}{2}\right) \left\| \tilde{g}_t \right\|^2 + \dfrac{\gamma}{2\rho} \left\| \nabla F(\hat{w}_t) - \nabla F(w_t) \right\|^2 \\
    &\hspace{2ex} + \left(\dfrac{\gamma \tau^2}{2\rho}+\frac{L \gamma^2 \tau^2}{2}\right)  \mathbb{E}_t \left[\left\|A_t\right\|^2\right] + \frac{L \gamma^2}{2} \sigma^2
\end{aligned}
$$    
$$
\begin{aligned}
    &\text{Since $F$ is $L-$smooth, we have:}  \\
    & \leqslant F\left(\hat{w}_t\right)-\gamma \left(1-\dfrac{\rho}{2}\right) \left\| \tilde{g}_t \right\|^2 + \dfrac{L^2\gamma}{2\rho} \left\| \hat{w}_t-{w}_t \right\|^2  \\
    & \hspace{5ex} + \left(\dfrac{\gamma \tau^2}{2\rho}+\frac{L \gamma^2 \tau^2}{2}\right)  \mathbb{E}_t \left[\left\|A_t\right\|^2\right] + \frac{L \gamma^2}{2} \sigma^2
\end{aligned}
$$    
$$
\begin{aligned}    
    &\text{Since $A_t = w_t-\hat{w}_t$, we have:}  \\
    & = F\left(\hat{w}_t\right)-\gamma \left(1-\dfrac{\rho}{2}\right) \left\| \tilde{g}_t \right\|^2  \\
    & \hspace{5ex} +\left(\dfrac{L^2\gamma}{2\rho} + \dfrac{\gamma \tau^2}{2\rho}+\frac{L \gamma^2 \tau^2}{2}\right)  \mathbb{E}_t \left[\left\|A_t\right\|^2\right] + \frac{L \gamma^2}{2} \sigma^2.    
\end{aligned}
$$
Using the bound for $\mathbb{E}_t \left[\left\|A_t\right\|^2\right]$ from Theorem~\ref{Theorem1} yields:
$$
\begin{aligned}
    &\mathbb{E}_t\left[F\left(\hat{w}_{t +1}\right)\right] \leqslant  F\left(\hat{w}_t\right)-\gamma \left(1-\dfrac{\rho}{2}\right) \left\| \tilde{g}_t \right\|^2  \\  
    &+\left(\dfrac{L^2\gamma}{2\rho} + r(\tau) \right)  \mathbb{E}_t \left[\left\|A_t\right\|^2\right] + \frac{L \gamma^2}{2} \sigma^2 \notag \\ 
    &\leqslant  F\left(\hat{w}_t\right)-\gamma \left(1-\dfrac{\rho}{2}\right) \left\| \tilde{g}_t \right\|^2  \\
    &+\frac{L \gamma^2}{2} \sigma^2  + \left(\dfrac{L^2\gamma}{2\rho} + r(\tau) \right)  \dfrac{4 \gamma^2 (1-\delta)\sigma^2}{\delta^2} \cdot \dfrac{1}{1+p(\tau)}. \notag\\
    \notag
\end{aligned}
$$
Rearranging terms yields:
$$
\begin{aligned}
    &\gamma \left(1-\dfrac{\rho}{2}\right) \left\| \tilde{g}_t \right\|^2  \leqslant  F\left(\hat{w}_t\right) - \mathbb{E}_t\left[F\left(\hat{w}_{t +1}\right)\right] + \frac{L \gamma^2}{2} \sigma^2  \\
    & \hspace{10ex} + \left(\dfrac{L^2\gamma}{2\rho} + r(\tau) \right)  \dfrac{4 \gamma^2 (1-\delta)\sigma^2}{\delta^2} \cdot \dfrac{1}{1+p(\tau)}. 
    \notag
\end{aligned}
$$
Then, averaging over $T$ iterations yields:
$$
\begin{aligned}
    \frac{1}{T+1} \sum_{t=0}^T\left\|\tilde{g}_t\right\|^2 
\end{aligned}
$$
$$
\begin{aligned}
    &\leq \dfrac{1}{\gamma\left(1-\frac{\rho}{2}\right)(T+1)} \sum_{t=0}^T\left(\mathbb{E}\left[F(\hat{w}_t\right)]-\mathbb{E}\left[F\left(\hat{w}_{t+1}\right)\right]\right) \notag\\
    &\hspace{3ex} \hspace{3cm}+\dfrac{L \gamma \sigma^2}{2\left(1-\frac{\rho}{2} \right)} \\
    &+\dfrac{1}{\gamma\left(1-\frac{\rho}{2} \right)} \left(\dfrac{L^2\gamma}{2\rho} + r(\tau) \right)  \dfrac{4 \gamma^2 (1-\delta)\sigma^2}{\delta^2} \cdot \dfrac{1}{1+p(\tau)}. 
    \notag
\end{aligned}
$$
As $T$ approaches infinity, since $\hat{w}_{t+1} =\hat{w}_t-\gamma \tilde{g}_t$, we have $\hat{w}_t$ approaching to the uncompressed SGD solution for $\tilde{F}(w)$
\begin{align}
    & \frac{1}{T+1} \sum_{t=0}^T\left\|\tilde{g}_t\right\|^2  \leq \dfrac{2(F(w_0) - F(w^*))}{\gamma\left(2-\rho\right)(T+1)} \notag  \\
    & +\dfrac{2}{\gamma\left(2-\rho \right)}  \left(\dfrac{L^2\gamma}{2\rho} + r(\tau) \right)  \dfrac{4 \gamma^2 (1-\delta)\sigma^2}{\delta^2} \cdot \dfrac{1}{1+p(\tau)}. 
    \notag
\end{align}
Setting $\rho=1$ completes the proof.
\noindent
By plugin in $\gamma =\dfrac{1}{\sqrt{T+1}} $, and using the definition of $r(\tau)$, we have:
$$
\begin{aligned}
& E_t ||\nabla \tilde{F}(w_{t+1})||^2   \le  \dfrac{2(F(w_0) - F(w^{*}))+L\sigma^2}{\sqrt{T+1}}  \\
&\hspace{10ex} +\left(    L^2 + 2 \sqrt{T+1} \cdot r(\tau) \right) \dfrac{4 (1-\delta) \sigma^2}{\delta^2 (T+1)} \cdot \dfrac{1}{1+p(\tau)}  \\ \notag
& =  \dfrac{2(F(w_0) - F(w^{*}))+L\sigma^2}{\sqrt{T+1}}  \\
& \hspace{10ex} +\left(    L^2 + \tau^2 + \dfrac{2\tau^2}{\sqrt{T+1}} \right) \dfrac{4 (1-\delta) \sigma^2}{\delta^2 (T+1)} \cdot \dfrac{1}{1+p(\tau)}  \\ \notag
& =  \dfrac{2(F(w_0) - F(w^{*}))+L\sigma^2}{\sqrt{T+1}} + L^2   \dfrac{4 (1-\delta) \sigma^2}{\delta^2 (T+1)} \cdot \dfrac{1}{1+p(\tau)}   \\
& \hspace{0ex} +\dfrac{4 (1-\delta) \sigma^2}{(\nicefrac{\delta}{\tau})^2 (T+1)} \cdot \dfrac{1}{1+p(\tau)}  \hspace{0ex} +\dfrac{2}{(\nicefrac{\delta}{\tau})^2} \dfrac{4 (1-\delta) \sigma^2}{\sqrt{T+1} (T+1)} \cdot \dfrac{1}{1+p(\tau)}, \notag
\end{aligned}
$$
which completes the proof. \hfill $\square$

\bibliographystyle{ieeetr}
\bibliography{bibliography}


\end{document}